\documentclass{libs/egpubl}

\JournalSubmission    
\CGFStandardLicense

\usepackage[T1]{fontenc}
\usepackage{libs/dfadobe}  
\usepackage{amsmath,amsfonts}
\usepackage{algorithmic}
\usepackage{algorithm}
\ifpdf \usepackage[pdftex]{graphicx} \pdfcompresslevel=9
\else \usepackage[dvips]{graphicx} \fi
\usepackage[utf8]{inputenc}

\usepackage{booktabs}
\usepackage{multirow}
\usepackage{adjustbox}
\usepackage{xcolor}
\usepackage{rotating}
\usepackage{fontawesome5}

\makeatletter
\AtBeginDocument{\def\@cite#1{[#1]}}
\makeatother

\newcommand{\rz}[1]{\textcolor{black}{#1}}
\newcommand{\mz}[1]{\textcolor{black}{#1}}

\newcommand{\saura}[1]{\textcolor{black}{#1}}

\newcommand{\snc}[1]{\textcolor{black}{#1}}
\newcommand{\mingz}[1]{\textcolor{black}{#1}}

\newcommand{\mrev}[1]{\textcolor{black}{#1}}
\newcommand{\gmark}[1]{\textcolor{black}{#1}}
\definecolor{taskNVS}{HTML}{2E5C9E}      
\definecolor{taskGEN}{HTML}{7E3FA0}      
\definecolor{taskART}{HTML}{D87A18}      
\definecolor{taskHUMAN}{HTML}{C2336B}    
\definecolor{taskLARGE}{HTML}{2E8B45}    
\definecolor{taskFEED}{HTML}{D4A017}     
\definecolor{taskPHYS}{HTML}{B53030}     

\newcommand{\iNVS}{\textcolor{taskNVS}{\faCamera}}
\newcommand{\iGEN}{\textcolor{taskGEN}{\faMagic}}
\newcommand{\iART}{\textcolor{taskART}{\faCogs}}
\newcommand{\iHUM}{\textcolor{taskHUMAN}{\faChild}}
\newcommand{\iLRG}{\textcolor{taskLARGE}{\faGlobe}}
\newcommand{\iFFW}{\textcolor{taskFEED}{\faBolt}}
\newcommand{\iPHY}{\textcolor{taskPHYS}{\faAtom}}

\definecolor{intPose}{HTML}{2E5C9E}   
\definecolor{intCont}{HTML}{2E8B45}   
\definecolor{intAff}{HTML}{D4A017}    
\definecolor{intPhys}{HTML}{B53030}   
\definecolor{intNeu}{HTML}{0E9594}    
\definecolor{intStr}{HTML}{555555}    

\newcommand{\iPose}{\textcolor{intPose}{\faStreetView}}
\newcommand{\iCont}{\textcolor{intCont}{\faHandPaper}}
\newcommand{\iAfd}{\textcolor{intAff}{\faBullseye}}
\newcommand{\iPReg}{\textcolor{intPhys}{\faWeightHanging}}

\BibtexOrBiblatex
\electronicVersion
\PrintedOrElectronic
\usepackage{cleveref}
\usepackage{array}
\usepackage[caption=false,font=normalsize,labelfont=sf,textfont=sf]{subfig}
\usepackage{textcomp}
\usepackage{stfloats}
\usepackage{url}
\usepackage{verbatim}
\usepackage{soul}
\usepackage{pifont}
\usepackage{booktabs}
\usepackage{multirow}
\usepackage[table]{xcolor}
\usepackage{pifont}
\usepackage[most]{tcolorbox}
\usepackage{lipsum}
\usepackage{libs/egweblnk} 
\usepackage{graphicx}

\newcommand{\am}[1]{\textcolor{black}{#1}}

\usepackage[normalem]{ulem} 

\electronicVersion

\title[Advances in 4D Representation: Geometry, Motion, and Interaction]%
      {Advances in 4D Representation: Geometry, Motion, and Interaction}

\author[M. Zhao et al.]
{\parbox{\textwidth}{\centering 
M. Zhao$^{1}$,
S. Nag$^{1}$, 
K. Wang$^{1}$,
A. Vora$^{1}$, 
G. Ji$^{1}$, 
P. Chun$^{2}$,
A. Mahdavi-Amiri$^{1}$, 
H. Zhang$^{1}$
}
\\
{\parbox{\textwidth}{\centering $^1$Simon Fraser University, $^2$ University of Alberta}
}
}

\begin{document}

\teaser{
 \includegraphics[width=0.9\linewidth]{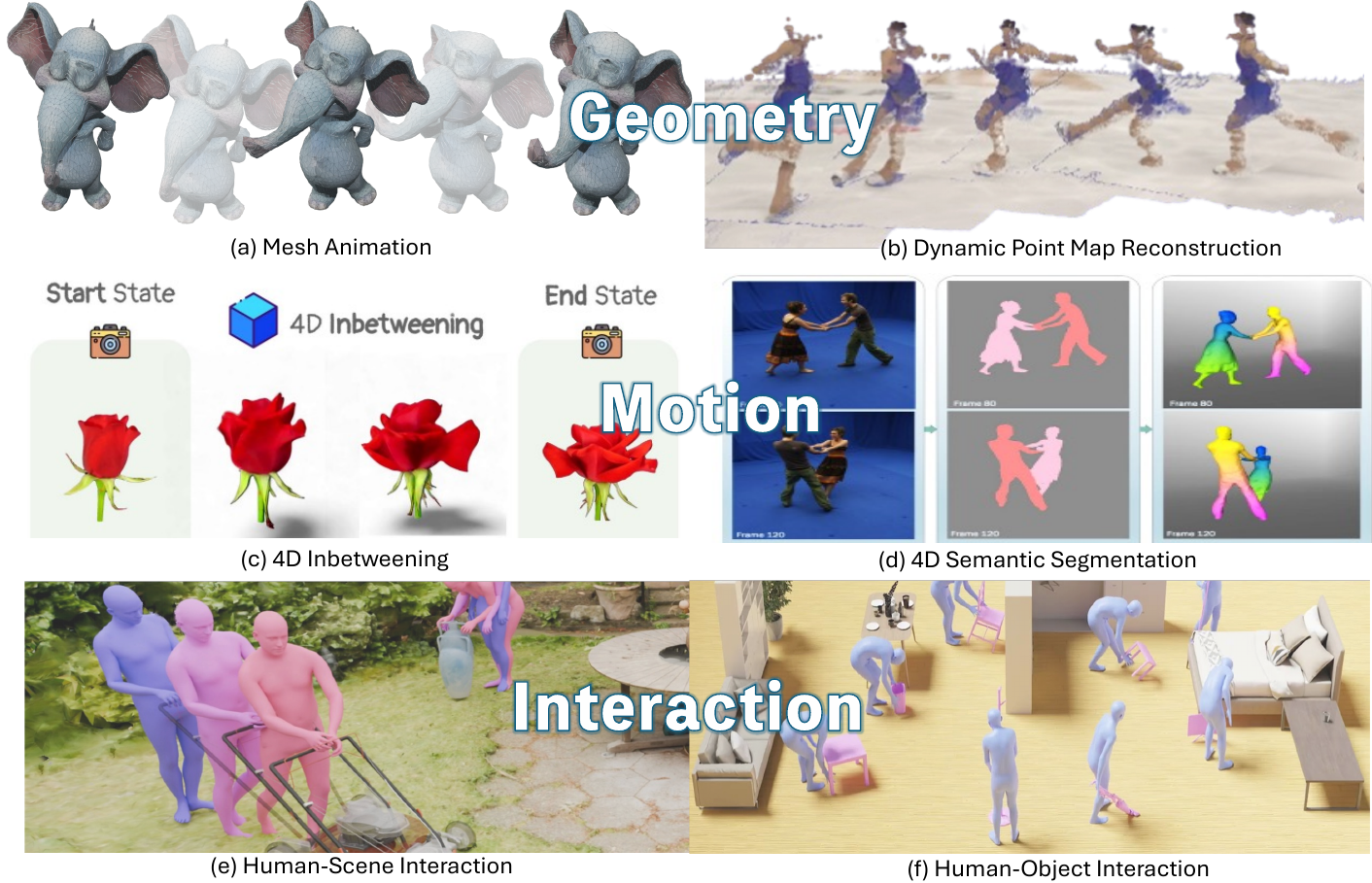}
 \centering
  \caption{
  Representative applications for three key pillars of \textbf{4D Representation}:
  (a) (Geometry) mesh animation~\cite{chen2025v2m4}, (b) (Geometry) dynamic point map reconstruction~\cite{wang2025pi}, (c) (Motion) 4D inbetweening~\cite{nag20252},(d) (Motion) 4D semantic segmentation \cite{mustafa20224d}, (e) Human-scene interaction~\cite{li2024zerohsi},(f) Human-object interaction~\cite{li2023object}.}
\label{fig:teaser}
}

\maketitle
\begin{abstract}



We present a survey on 4D generation and reconstruction, a fast-evolving subfield of computer graphics whose developments have been propelled by recent advances in neural fields, geometric and motion deep learning, as well as 3D generative artificial intelligence (GenAI). While our survey is not the first of its kind, we build our coverage of the domain from a unique and distinctive perspective of 4D {\em representations\/}, to model 3D geometry evolving over time while exhibiting motion and interaction. Specifically, \gmark{instead of offering an exhaustive enumeration of many works, we take a more selective approach by focusing on representative works to highlight both the desirable properties and ensuing challenges of each representation under different computation, application, and data scenarios.} The main take-away message we aim to convey to the readers is on how to select and then customize the appropriate 4D representations for their tasks. Organizationally, we separate the 4D representations based on three key pillars: {\em geometry\/}, {\em motion\/}, and {\em interaction\/}. Our discourse will not only encompass the most popular representations of today, such as neural radiance fields (NeRFs) and 3D Gaussian Splatting (3DGS), but also bring attention to relatively under-explored representations in the 4D context, such as {\em structured\/} models and long-range motions. Throughout our survey, we will reprise the role of large language models (LLMs) and video foundational models (VFMs) in a variety of 4D applications, while steering our discussion towards their current limitations and how they can be addressed. We also provide a dedicated coverage on what 4D datasets are currently available, as well as what is lacking, in driving the subfield forward. 

\end{abstract}  

\section{Introduction}
\label{sec:intro}

Reconstruction and generation of 4D data, i.e., 3D geometry evolving over time while exhibiting motion and interaction, is becoming increasingly critical as computer graphics applications expand into domains requiring dynamic scene understanding, temporal modeling, and motion synthesis. From cinematic visual effects and immersive virtual reality to autonomous robotics, medical imaging, eCommerce and advertising, the ability to capture, represent, and manipulate 4D content has emerged as a fundamental challenge that bridges graphics, vision, and machine learning.

The fourth dimension introduces complexities that extend far beyond simply concatenating spatial coordinates with temporal indices. Temporal coherence, motion continuity, topological changes, interaction dynamics, and the preservation of geometric, appearance, and physical properties across time present unique {\em representational\/} challenges that require careful consideration of both spatial and temporal encoding strategies. As the field matures, researchers have developed increasingly sophisticated approaches to handle these challenges, leading to a rich landscape of 4D representation schemes, each with distinct advantages and limitations.

Recent surveys in this domain have primarily categorized methods by applications, approaches, and/or the granularity of the extracted scene signals. 
Cao et al.~\cite{cao2025reconstructing4dspatialintelligence} provides a structured review of recent progress in reconstructing 4D spatial intelligence, which is defined as {\em understanding\/} 3D scenes and their evolution over time, advancing from extracting low-level geometric cues such as depth, pose, and point maps toward complex reasoning encompassing interactions and physics. This survey does highlight the growing importance of advanced 3D representations while focusing mainly on neural radiance fields (NeRFs) and
3D Gaussian splatting (3DGS). 
Along the same lines, Fan et al.~\cite{fan2025advancesradiancefielddynamic} presents a comprehensive survey of recent progress in dynamic scene representation and reconstruction, focusing on the transition from NeRF to 3DGS and their trade-offs. This second survey systematically categorizes the works covered based on motion representation paradigms including rigid, articulated, non-rigid, and hybrid motions, and examines strategies for modeling temporal changes in geometry and appearance.
In addition, Miao et al.~\cite{Miao2025Advances} survey the emerging field of 4D {\em generation\/} by introducing a taxonomy of low-level geometry representations (meshes, NeRFs, point clouds, and 3DGS), foundational techniques (diffusion models and score distillation sampling), and conditioning methods (text, images, videos, 3D inputs, and multimodal control). This survey further categorizes the algorithmic approaches (end-to-end, via generated data, implicit distillation, vs.~explicit supervision) and highlights the growing range of applications for 4D generation.

While the above surveys have advanced our understanding of the field's breadth, they have not sufficiently covered all the relevant representations of geometry (especially higher-level {\em structured\/} representations), motion, and interaction, or addressed the fundamental question of why specific representations are chosen for particular 4D tasks and the motion/interaction mechanisms, nor have they provided comprehensive analysis of the inherent trade-offs that different representational choices impose on data preparation, method design, computational requirement, and achievable results.

To close these gaps, our survey adopts a {\em representation-centric\/} perspective. 
\gmark{Instead of offering an exhaustive enumeration of many works, we take a more selective approach by focusing on representative works to highlight both the desirable properties and ensuing challenges of each 4D representation under different computation, application, and data scenarios.}
\mrev{Our curation rests on three criteria: representational novelty (methods that introduce or substantially extend a 4D representation), influence and recency (foundational works alongside recent state of the art), and pillar coverage (works that collectively span the geometry, motion, and interaction pillars of our survey).}
The main take-away message we aim to convey to the readers is on how to select and then customize the appropriate 4D representations for their tasks. 

Organizationally, we separate the 4D representations based on three key pillars: {\em geometry\/}, {\em motion\/}, and {\em interaction\/}. We distinguish between structured representations, i.e., those that maintain explicit primitive or part delineations and relations, and unstructured representations, i.e., those that encode 4D content through implicit functions, point distributions, or learned feature spaces without explicit structural constraints. While the latter category encompasses the most popular representations of today, namely NeRFs and 3DGS, they are both built on {\em rendering\/} primitives, mainly for novel view synthesis. They are not best suited to prevalent 4D tasks such as modeling, editing, or interactions, for which more compact and structural representations, built on higher-level primitives, are more appropriate. Hence, we also discuss these representational choices even though they have been relatively under-explored. Throughout our survey, we will reprise the role of large language models (LLMs) and video foundational models (VFMs) in a variety of 4D applications, while steering our discussion towards their current limitations and how they can be addressed. We also provide a dedicated coverage on what 4D datasets are currently available, as well as what is lacking, in driving the subfield forward.

Our representation-centric analysis goes beyond traditional surveys in several key ways. First, we examine not only the capabilities of each representation but also their fundamental limitations and the specific 4D challenges they are designed to address. Second, we analyze how representational choices constrain and enable different architectural approaches for learning, predicting, and generating temporal dynamics. Third, we provide comprehensive coverage of motion modeling approaches, temporal consistency mechanisms, and the interplay between representation and motion characteristics aspects that have received limited attention in previous surveys despite their central importance to 4D applications.

Furthermore, this survey addresses the growing importance of motion analysis in 4D representation by dedicating significant attention to temporal dynamics, motion types, and their representational requirements. We examine how different motion characteristics, from rigid articulations to non-linear deformations to topological changes, interact with representational choices and impose constraints on method design. This motion-centric analysis is complemented by comprehensive coverage of datasets, evaluation metrics, and benchmarking protocols that have emerged to support systematic comparison of 4D methods across different representations.

\begin{figure*}
    \centering
    \includegraphics[width=\linewidth]{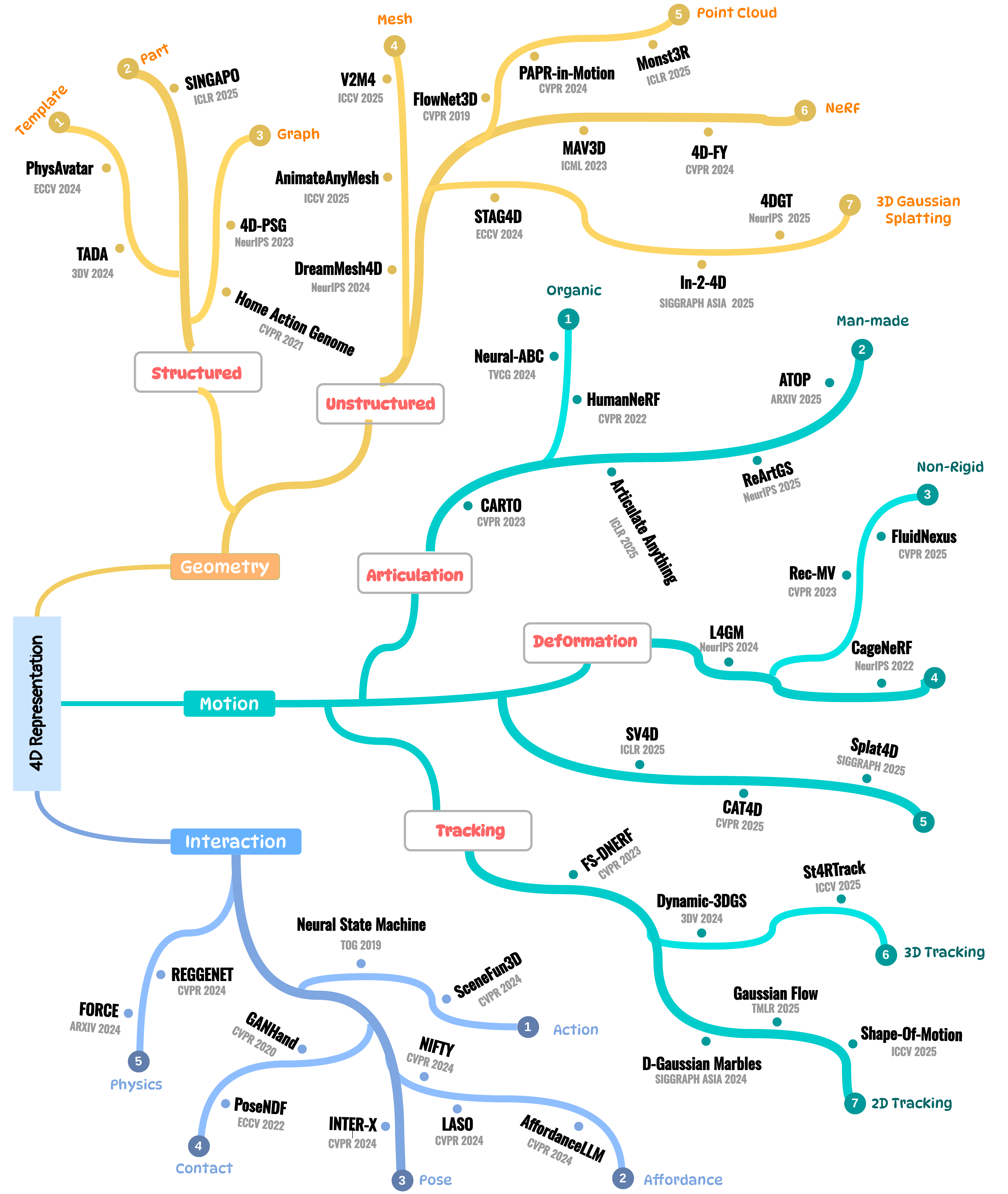}
    \caption{Our taxonomy of 4D representations. We separate them into three pillars: (1) Geometry, including both structured and unstructured representations; (2) Motion, including articulation, deformation and tracking based representations; (3) Interaction, including representation of action, affordance, pose, contact and physics.}
    \label{fig:taxonomy}
\end{figure*}

Our survey is organized as follows. \Cref{sec: geometry} presents our core representational analysis, examining unstructured representations (mesh, point clouds, NeRFs and 3DGS) and structured representations (template,  parts, and graphs). \Cref{sec: motion} focuses on motion and temporal dynamics, analyzing how different motion types interact with representational choices. \Cref{sec:interac} covers representation choices in modeling multiple entities that interact with each other. \Cref{sec: dataset} surveys datasets, evaluation metrics, and benchmarking frameworks that enable systematic comparison across representations. \Cref{sec: training paradigm} provides an overview on training paradigms. Finally, \Cref{sec:rep_tradeoff,sec: Conclusion} provide an overall analysis across representations and discuss emerging trends, persistent challenges, and future research directions in 4D representation. The overall taxonomy of our survey is shown in \Cref{fig:taxonomy}.

\begin{figure*}
    \includegraphics[width=\linewidth]{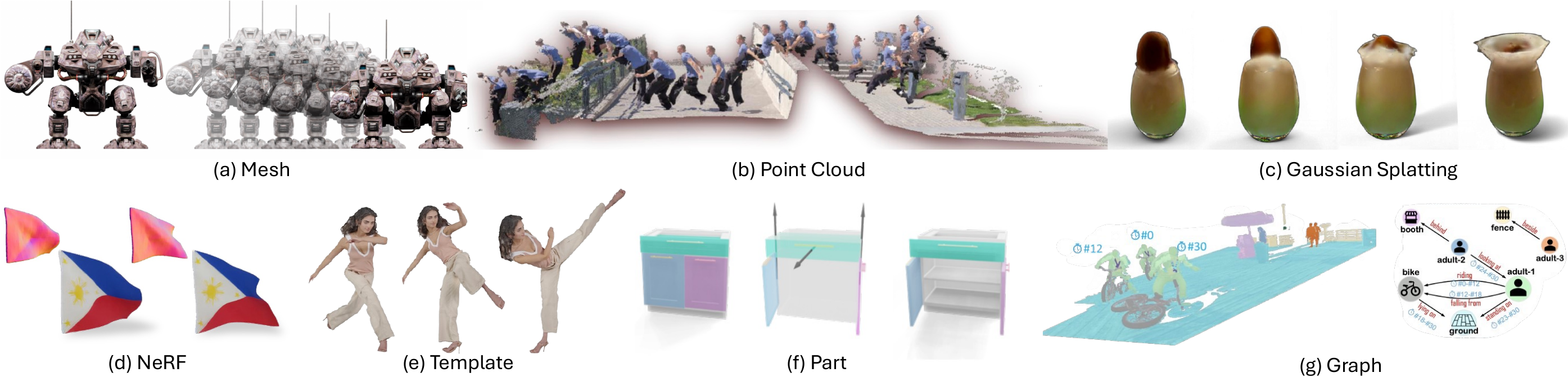}
    \caption{4D content created with different geometric representations. (a) Mesh~\cite{wu2025animateanymesh}; (b) Point cloud~\cite{wang2025pi}; (c) Gaussian Splatting~\cite{nag20252}; (d) NeRF~\cite{voleti2024sv3d}; (e) Template-based representation~\cite{PhysAavatar24}; (f)  Part-based representation~\cite{liu2024cage}; (g) Spatial-Temporal scene graph~\cite{4DPanoSceneGraph}. Figures adopted from the original papers.}
    \label{fig: geometry}
\end{figure*}

\section{Modeling Geometry}
\label{sec: geometry}
\am{Each 3D representation for 4D carries advantages and disadvantages.} 
We survey geometric representations in state-of-the-art 4D works, examining not only their intrinsic properties but also their \am{integration} with the current 4D tasks: which representations are favored for particular applications, what unique capabilities they enable, and what \am{limitations they may have.}

We categorize representations into \textit{unstructured} (\Cref{sec: unstructured representation}) and \textit{structured} (\Cref{sec: structured representation}) classes based on whether their operational primitives carry explicit functional, hierarchical, or semantic meaning. See \Cref{fig: geometry} for an illustration.
Unstructured representations—meshes (\Cref{subsec: mesh rep}), point clouds (\Cref{subsec: pcd rep}), NeRF (\Cref{subsec: nerf}), and Gaussian Splatting (\Cref{subsec: gs})—use primitives (vertices, points, ray samples, Gaussians) as independent geometric elements. While local connectivity may exist (e.g., mesh faces), these representations do not impose inter-primitive global structural constraints or functional decompositions.

Structured representations—skeletal templates (\Cref{subsec: template-based representation}), part-based models (\Cref{subsec: primitive representation}), and graphs (\Cref{subsec: graph based representation})—impose functional decomposition: skeletal joints define kinematic chains, parts carry semantics, graph nodes encode relational structure. 
\am{Some} structured representations build on unstructured primitives (e.g., SMPL uses a mesh but adds skeletal hierarchy and skinning weights); the key distinction lies in whether functional relationships and compositional structure are explicitly encoded as first-class constraints in the representation.

\subsection{Unstructured Representation}
\label{sec: unstructured representation}
Unstructured representations form the foundation of most 4D reconstruction and generation pipelines due to their flexibility and widespread adoption in graphics and vision. We examine four dominant paradigms—meshes, point clouds, NeRF, and Gaussian Splatting—analyzing how each adapts to temporal dynamics and the specific challenges they face in 4D settings.

\subsubsection{Mesh-Based Representations}
\label{subsec: mesh rep}
A polygonal mesh represents a 3D object's surface through vertices (points in 3D space), edges (connecting vertex pairs), and faces (surfaces enclosed by edge loops), explicitly encoding both geometry through vertex coordinates and topology through connectivity relationships. 
Meshes retain \am{important} advantages in graphics applications: compatibility with GPU rasterization enables real-time rendering; 
explicit structure allows intuitive manipulation and editing; natural decoupling of geometry, topology, and appearance enables motion generation through vertex displacement while maintaining constant topology and texture maps across frames, ensuring temporal coherence and object identity preservation.

Meshes are mainly adopted in object-centric 4D modeling, aimed at producing animated mesh assets from text~\cite{TextMesh4D,chen2024ct4d}, video~\cite{chen2025v2m4,li2024dreammesh4d}, or static 3D assets~\cite{wu2025animateanymesh,shi2025drive}. However, mesh representations face several challenges in 4D applications, which we discuss below alongside methods developed to address them.

\noindent\textbf{Spatiotemporal Consistency:} Animated mesh sequences require precise vertex-to-vertex correspondence across frames to preserve consistent texture and topology, preventing naive per-frame generation pipelines that can lead to different \am{vertex-connectivity} in each generated sample. V2M4~\cite{chen2025v2m4} circumvents this through a reconstruction-and-refine pipeline: it first applies image-to-3D reconstruction models~\cite{xiang2025structured,zhao2025hunyuan3d,li2025triposg,li2024craftsman} to generate per-frame meshes, then sets the first frame as an anchor and deforms it to register with subsequent frames, maintaining consistent mesh correspondence through continuous deformation.
    
\noindent\textbf{Deformation Learning:} Given that most approaches rely on deforming a consistent mesh topology rather than per-frame generation, determining correct vertex displacements becomes the central challenge. Vertices lack inherent semantic meaning, making it difficult to learn meaningful deformation patterns. Two primary approaches have emerged: directly learning per-vertex displacement from data priors~\cite{wu2025animateanymesh} or video diffusion model guidance~\cite{TextMesh4D}, or introducing an intermediate rigging structure—either explicit~\cite{Song_2025_CVPR,song2025puppeteer} or approximated~\cite{li2024dreammesh4d}—followed by skinning-based articulation to propagate skeletal motion to vertex displacements.

\noindent\textbf{Open Challenges:} Despite recent progress, fundamental limitations remain. Topological inflexibility prevents meshes from representing splitting or merging. Their \am{inflexible topology and connectivity} struggle with volumetric phenomena such as clouds, smoke, and fire, motivating the adoption of topologically flexible alternatives such as NeRF~(\Cref{subsec: nerf}) and point clouds~(\Cref{subsec: pcd rep}) for such scenarios. Additionally, current mesh-centric generation and reconstruction methods predominantly focus on object-level rather than scene-level modeling. The latter has recently been studied on static 3D scene reconstruction~\cite{yao2025cast} through generative methods, while most existing works rely on mesh database retrieval~\cite{wu2024diorama}. Key difficulties include: (1) generating and tracking separate mesh topologies for multiple objects at varying scales, (2) handling inter-mesh physical constraints, and (3) maintaining temporal coherence when objects move independently with different deformation patterns. Addressing these challenges remains an open problem for the research community.

\subsubsection{Point-cloud-based Representation}
\label{subsec: pcd rep}

A point cloud represents 3D geometry as an unordered set of points $P = \{p_i\}_{i=1}^N$, where each point $p_i$ is defined by spatial coordinates $(x, y, z)$ and optional attributes including color (RGB), intensity, surface normals, and timestamps. Their unstructured, unordered nature enables representation of arbitrary and changing topologies without connectivity constraints. Point clouds also serve as the primary output format for real-world acquisition \am{mechanisms}. This direct sensor correspondence has established point clouds as the standard representation for large-scale scene capture, particularly in autonomous driving (KITTI~\cite{geiger2013vision}, nuScenes~\cite{caesar2020nuscenes}), where their geometric scalability outperforms alternatives. \mz{On native 4D point cloud sequences, numerous works focus on improving point cloud quality through denoising~\cite{hu2025dynamic}, upsampling~\cite{li2021tpu,berlincioni20234dsr}, and mesh reconstruction~\cite{tavu2022rfnet4d,rempe2020caspr,niemeyer2019occupancy}. Another research direction addresses dynamic point cloud understanding, including motion recognition~\cite{liu2025mamba4d,dong2023nsm4d}, object detection and tracking~\cite{huang2022dynamic,wei2024t,cao2024motion2vecsets}, scene flow estimation~\cite{liu2019flownet3d,wu2020pointpwc,teed2020raft}, forecasting~\cite{khurana2023point,wang2024semantic}, and motion interpolation~\cite{zeng2022ideanet,zheng2023neuralpci,lu2021pointinet}. With the emergence of large feed-forward geometric models like DUST3R~\cite{wang2024dust3r} and VGGT~\cite{wang2025vggt}, a new promising branch of work is to create dynamic point map sequences from videos~\cite{zhang2024monst3r,cut3r,wang2025pi,team2025aether} and incorporate tracking within the inferred point maps~\cite{xiao2025spatialtrackerv2}.}

Below we discuss the main challenges in applying point clouds to 4D contexts and representative works addressing them.
    
  \noindent \textbf{Temporal Correspondence:} Different frames in dynamic point cloud sequences exhibit varying noise patterns, point counts, and no guaranteed point-to-point correspondence. FlowNet3D~\cite{liu2019flownet3d} pioneered end-to-end scene flow learning, estimating vector displacements between consecutive frames without assuming point-to-point correspondence. St4RTrack~\cite{feng2025st4rtrack} generates dynamic point cloud sequences from videos and tracks point motion using an attention-enabled tracking branch that leverages video inputs. Alternatively, correspondence can be enforced by interpolating between adjacent frames. FastPCI~\cite{zhang2024fastpci} applies pyramidal transformer blocks to extract motion and structure features, facilitating in-betweening generation.
    
\noindent \textbf{Appearance Modeling:} The discrete, isolated nature of point clouds prevents complete appearance modeling, leaving void spaces between points unlike surface or volumetric representations. PAPR-in-Motion~\cite{peng2024papr} addresses this by associating each point with view-dependent features and rendering through attention-weighted feature interpolation, achieving void-free rendering while enabling dynamic visualization through point displacement.

\noindent\textbf{Open Challenges:} Point clouds remain the primary interface with real-world sensor captures (LiDAR, RGB-D cameras), and this abundance of point-based training data has recently enabled pointmap-based methods to achieve impressive 4D reconstruction from monocular videos. However, a reconstruction-generation asymmetry persists: while point clouds excel at capturing dynamic geometry from 
observations, their inherent limitations—discrete sampling without explicit surface connectivity—make them less suitable for generation tasks. Incorporating physical motion remains challenging: the 
lack of topological structure complicates enforcement of physical constraints such as collision response, conservation laws, and contact dynamics. Future work should focus on efficient bidirectional conversion between point clouds and other representations, enabling knowledge transfer of mesh topology, NeRF appearance modeling, and physical constraints to point-based frameworks while preserving their computational efficiency and sensor correspondence advantages.

\subsubsection{Neural Radiance Fields (NeRF) based  Representation}
\label{subsec: nerf}
Neural Radiance Fields (NeRF)~\cite{mildenhall2021nerf} represent 3D scenes by mapping 3D query points and view directions to density and color using an MLP. Final pixel colors are obtained by integrating these color predictions along camera rays via volume rendering~\cite{kajiya1984ray}. This continuous representation achieves high visual fidelity from only 2D images and can capture topologically complex phenomena like smoke and fluids, and compactly store dynamic scenes within an MLP~\cite{park2021hypernerf,attal2023hyperreel}. Despite its promise, NeRF faces several key challenges in 4D modeling.

\noindent \textbf{Flickering and Unrealistic Deformations:} A direct extension of NeRF to the 4D domain models temporal variations by introducing time-varying fields. This is typically done by conditioning the representation on a time signal or by learning a unique embedding for each timestamp ~\cite{gao2021dynamic, li2022neural, wu2022d, ost2021neural}. Another approach models dynamics by predicting deformation or flow fields using a separate deformation network, which warps the scene over time ~\cite{li2021neural, song2023nerfplayer}. Each frame’s color and density are then composited and rendered using the standard NeRF volume rendering framework. However, this approach often suffers from temporal flickering and unrealistic deformations. These issues are typically alleviated through temporal consistency losses and rigidity constraints~\cite{gao2021dynamic, li2021neural}.

\noindent \textbf{Slow Convergence:}
NeRF-based representations often suffer from slow convergence due to the heavy computational cost of training large MLPs for dynamic scenes. To accelerate training, several methods introduce learnable spatial-temporal data structures such as Hex-planes and K-planes~\cite{cao2023hexplane, singer2023text, fridovich2023k, jiang2023consistent4d}, which trade memory for speed. These approaches sample positional features from planar decompositions via bilinear interpolation before passing them through a smaller MLP for color and density prediction. By offloading computation from the MLP to explicit plane-based storage, convergence accelerates significantly compared to standard NeRFs. However, planar decompositions may sacrifice fine-grained spatial details. Alternative data structures have also been explored, including voxel grids~\cite{liu2022devrf, fang2022fast, shao2023tensor4d, yang2023real} that provide denser spatial sampling, and hash grids~\cite{bahmani20244dfy, zhao2023animate124, zhang_4diffusion_2024} that offer adaptive resolution through learned hash encodings.

\noindent \textbf{Handling Sparse Input Scenarios:} NeRF representations, in general, require dense multi-view video captures as input. To improve robustness under sparse input conditions, recent works incorporate flow supervision~\cite{wang2023flow, yao2024neural} to train models with only monocular video supervision. Other methods approach this problem through motion-adjusted feature aggregation~\cite{li2023dynibar} or enriching input data from generative models~\cite{yao2025sv4d, xie_sv4d_2024}.

\noindent\textbf{Open Challenges:} Despite these improvements, temporal flickering and unrealistic motion artifacts remain persistent issues, particularly when training from sparse or limited input views. Motion synthesis and editing in NeRF-based methods are also less intuitive, as manipulations occur in latent deformation spaces rather than through direct geometric controls. Additionally, NeRF's ray marching mechanism incurs significant computational overhead during inference, leading the community to increasingly favor Gaussian Splatting(\Cref{subsec: gs}) as a more efficient alternative for real-time dynamic scene rendering.

\subsubsection{Gaussian Splatting based Representations}
\label{subsec: gs}
3D Gaussian Splatting~\cite{kerbl3dgaussians} represents 3D scenes using a set of Gaussian primitives and renders images through rasterization. Each Gaussian is defined by its 3D covariance matrix and spatial position, and is optimized using photometric loss after rasterization. This explicit, discrete representation offers several advantages: it avoids redundant computation in empty regions by only evaluating non-zero Gaussians, enables real-time differentiable rendering through efficient tile-based rasterization~\cite{kerbl3dgaussians}, and supports motion modeling more explicitly through deforming each Gaussian. Gaussian Splatting's explicit formulation and efficient rendering pipeline have made it increasingly favored for dynamic scene applications. Despite these benefits, several challenges remain for this representation in 4D settings.

\noindent\textbf{Temporal Coherency:} A straightforward way to model dynamic scenes is to generate per-frame 3D Gaussians~\cite{ren2024l4gm, vora2025articulate} from an input video. However, maintaining temporal coherence across frames remains difficult due to flickering and blurriness caused by dynamically learned Gaussian colors. To mitigate this, many methods opt for continuous deformation field~\cite{zeng2024stag4d, yin_4dgen_2023,wu2025cat4d,yao2025sd,xu2024grid4d,shao2024control4d} or tracking-based deformation~\cite{stearns2024dynamic, lei2024mosca} or attention-based deformation~\cite{sun_eg4d_2024} to enforce temporal coherency through local regularization. Another emerging trend is to use native 4DGS that encodes space-time variations within the geometry representation~\cite{duan20244d,Wu_2024_CVPR}.

\noindent\textbf{Sparse Input Conditions:} Similar to NeRF, the most comprehensive input for Gaussian Splatting is also multi-view synchronized videos. However, such data scales drastically in volume with video length and number of view angles, along with the induced computational burden, such data is also harder to acquire. Thus it is a natural choice for the community to shift toward a sparse input paradigm. Existing methods include simultaneously learning canonical static Gaussian and deformation field from monocular videos~\cite{yang20244d}, leveraging additional physical priors such as optical flow~\cite{gao2024gaussianflow,lin2024gaussian}, depth~\cite{liu2025modgs} and video diffusion guidance~\cite{nag20252,ren2024l4gm,wu2025cat4d}.

\noindent \textbf{Open Challenges:} Improving spatial-temporal coherence, training efficiency, and adaptation to sparse inputs remain active research areas. Beyond these, modeling large-scale motion in Gaussian Splatting poses significant challenges, as it requires carefully calibrated covariance matrices to properly reorient and deform Gaussian primitives across frames. Additionally, while volumetric rendering methods inherently couple appearance and geometry, extracting high-quality explicit meshes from dynamic Gaussian representations remains an important open problem.

\subsection{Structured Representation}
\label{sec: structured representation}
Structured representations impose compositional priors that enable more controllable and interpretable 4D modeling. We survey three primary approaches—template-based, part-based, and graph-based methods—examining how explicit structural constraints facilitate motion modeling and the challenges they present.

\subsubsection{Template-based Representation}
\label{subsec: template-based representation}
Template-based representations combine a parametric mesh with an underlying skeletal structure (kinematic tree). Mesh vertices are bound to the skeleton through a \textit{skinning} function that determines vertex deformation during skeletal articulation. Animation occurs by manipulating skeletal pose parameters, which drive mesh deformation through skinning weights (detailed in \Cref{subsec:template_articulation}). This representation models object categories sharing common topology and articulated structure—humans, hands, or animals—while allowing individual variation in shape and pose. Examples include SMPL~\cite{SMPL:2015} for human bodies, MANO~\cite{MANO:SIGGRAPHASIA:2017} for hands, SMAL~\cite{Zuffi:CVPR:2017} for animals, FLAME~\cite{FLAME:SiggraphAsia2017} for heads, and compositional variants like SMPL+X~\cite{pavlakos2019expressive} for body, hand, and face.
Templates offer distinct advantages for 4D modeling. They inherently couple motion within parametric structure, provide compactness while maintaining anatomical plausibility, and encode category-specific priors enabling realistic motion and facilitating motion transfer. By decomposing 4D modeling into per-frame pose~\cite{goel2023humans}, templates effectively handle long-term sequences, establishing foundations for digital avatars, human-object interactions, and hand manipulation applications.
Despite these advantages, templates face challenges. 

\noindent\textbf{Sub-realism:} Models like SMPL represent bare bodies with rigid motion, leaving gaps in realistic digital human modeling. Enhanced personalization requires deforming meshes to mimic cloth~\cite{CAPE:CVPR:20}, texturing for appearance~\cite{liao2024tada}, or generating garments and hair with simulators~\cite{simavatar2024,PhysAavatar24}. SoftSMPL~\cite{santesteban2020softsmpl} introduces soft tissue dynamics as a preliminary step toward realism. 

\noindent\textbf{Temporal Consistency:} Per-frame pose estimation can produce temporal jitter from frame-level errors, resulting in unrealistic motion despite correct per-frame geometry. This can be addressed by temporal smoothness constraints through sequential models~\cite{kanazawa2019learning,kocabas2020vibe,choi2021beyond} that leverage temporal context from adjacent frames, or physics-based priors~\cite{vladislav2020physcap} that enforce motion continuity and physical plausibility. 

\noindent\textbf{Open Challenges:} While template-based representations exhibit excellent motion transferability within categories, cross-category transfer remains limited. Developing generalizable template learning frameworks that automatically infer skeletal structures, skinning weights, and shape spaces from sparse examples would greatly boost the representation's applicability~\cite{liu2025riganything,Song_2025_CVPR}. Furthermore, integrating physics simulation with learned skinning functions to enable realistic contact dynamics, collision response, and soft tissue deformation represents a promising direction.

\subsubsection{Part-based Representation}
\label{subsec: primitive representation}

Part-based representations decompose 3D entities into individual parts and model overall motion through their collective dynamics. Unlike unstructured representations with holistic motion models, part-based approaches assign semantic meaning to decomposed parts and functional meaning to their motion. Parts can be represented using various formats such as NeRF~\cite{liu2023paris}, point clouds~\cite{yan2020rpm}, or meshes~\cite{jiang2022opd}. This representation is commonly used for modeling articulated objects~\cite{liu2024cage,liu2024singapo,iliash2024s2o,lei2023nap} —such as cabinets with opening doors, pulling drawers, and rotating hinges—with applications in embodied AI and robotic manipulation.

Part-based representation faces two primary challenges: decomposing 3D entities into functional parts and correctly articulating each primitive. The decomposition process requires functional understanding of component relationships and involves both segmentation and completion during part reconstruction. Early approaches rely on segmented datasets~\cite{Xiang_2020_SAPIEN} but show limited generalizability due to dataset scale constraints. Recent work leverages large language models for semantic reasoning in part segmentation and completion~\cite{qiu2025articulate,liu2024singapo,xia2025drawerdigitalreconstructionarticulation,lu2025dreamart}. These methods use LLMs to identify functional parts and plausible articulated motions, then apply amodal completion on segmented 3D parts or retrieval-based reconstruction. An alternative approach learns kinematic part decomposition directly from visual data. SP4D~\cite{zhang2025sp4d} jointly generates multi-view RGB videos and corresponding kinematic part segmentation from monocular inputs. The method lifts the generated 2D part maps to 3D to derive skeletal structures and harmonic skinning weights, enabling articulated 3D asset creation for extended categories.

\noindent\textbf{Open Challenges:} Part-based representation remains a growing field. Active research continues in both 2D~\cite{li2024dragapart,li2024puppet,vora2025articulate,sun2024opdmulti,wang2024active} and 3D~\cite{zhao2024sweepnet,yu2024dpa,yang2025holopart,ye2025primitiveanything}, with many techniques yet to be transferred to 4D content creation. Precisely segmenting and articulating parts of 3D models remains an open problem, with current applications concentrated on specific object categories such as furniture and relatively simple motions—prismatic translation, rotation, and revolution. The underlying challenge still lies in functional understanding. Extending to more complex behaviors such as per-part non-rigid deformation and generic objects requires a deeper functional understanding of 3D assets and their constituent parts. These directions warrant further investigation by the research community.

\subsubsection{Spatio-Temporal Scene Graphs}
\label{subsec: graph based representation}
Scene graphs are a popular representation for describing multiple geometries in an environment and the relations between them.
3D scene graphs could be generalized to include a temporal dimension through multiple means.
A straightforward approach is to represent a 4D spatio-temporal scene graph as a collection of 3D scene graphs at different time stamps~\cite{ActionGenome, HOMAGE}.
While each slice is a standard 3D scene graph, temporal edges can be used to connect the same entity across slices.
It is also possible to generalize a linear sequence into an arbitrary graph, where edges link temporally proximal concepts~\cite{MotionGraphs}.
A collection of slices could also be grouped to represent entire events.

While such representations are effective, they are computationally expensive---when the size and length of the 4D scene grow, because all nodes are still in 3D.
A natural generalization is to adopt 4D nodes, where each node describes a tube/trajectory of 4D geometries over time~\cite{ICCV11STGraph}, rather than a single 3D geometry.
In addition to concrete geometries, 4D nodes can also represent abstract concepts such as events.
Doing so drastically reduces the size of the graph, removing the need for a node at every time slice.
Edges can also be generalized to work with 4D nodes.
Instead of distinguishing between spatial (at 3D slices) and temporal (connecting adjacent 3D nodes) edges, a single edge connecting 4D nodes can represent spatio-temporal concepts over time, including time-restricted properties (e.g. co-visibility)~\cite{3DDSG}, evolving actions (e.g. falling from)~\cite{4DPanoSceneGraph}, or causal relationships.

\noindent\textbf{Open Challenges:}
While 4D Spatio-Temporal scene graphs naturally describe structural relations and semantics better, they are more complex than the unstructured counterparts.
Curating compatible 4D data and developing specific learning algorithms for such representation also require significant efforts ~\cite{ICCV11STGraph, STGCN, InteractionNetworks}.
Consequently, achieving comparable geometric fidelity and temporal granularity to unstructured 4D representations remains challenging.
Connecting such 4D graphs as an additional layer of abstraction over aforementioned unstructured 4D representations is an important future research direction.
\section{Modeling Motion}
\label{sec: motion}

\begin{figure*}
    \centering
    \includegraphics[width=1\linewidth]{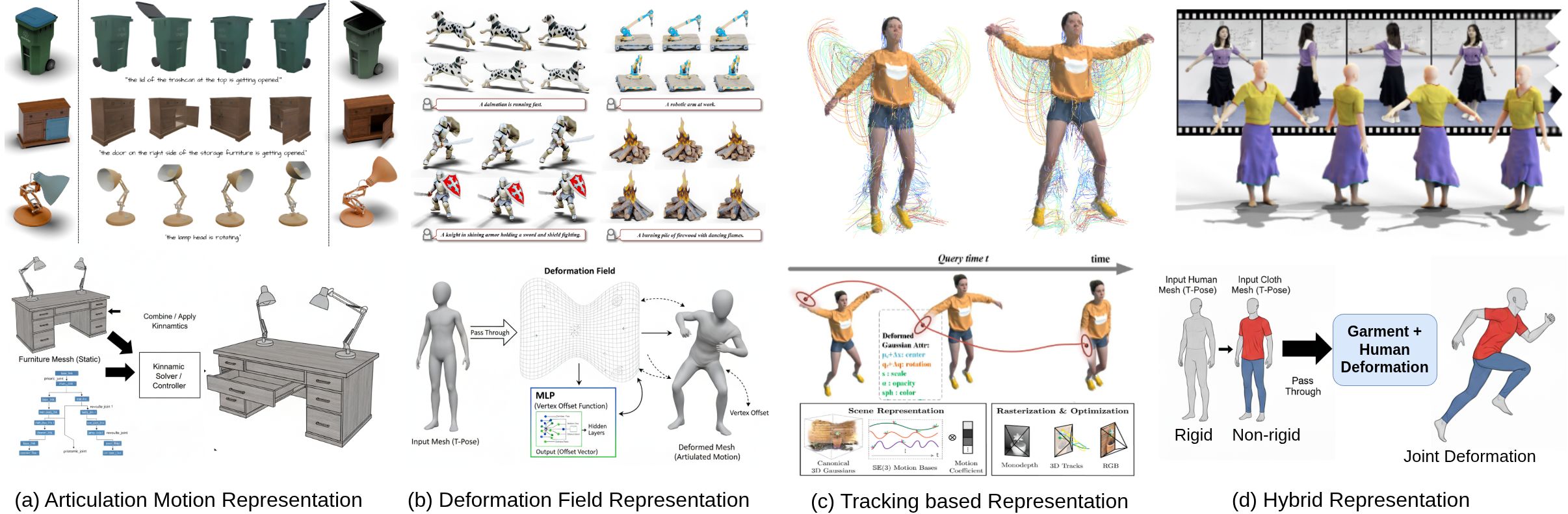}
    \caption{Various examples showing how motion is added to the geometry. From left to right:~\cite{vora2025articulate, TextMesh4D, luiten2024dynamic, qiu2023rec}}
    \label{fig:motion}
    \vspace{-0.5em}
\end{figure*}

The core of motion representation is establishing how geometry evolves across temporal frames. This can be formulated as determining the geometric state $\mathcal{G}_t$ as time $t$ evolves. \gmark{We distinguish four principal motion classes in 4D representations: articulated motion, deformation-based motion, tracking-based motion, and hybrid motion} (Fig.~\ref{fig:motion}), discussed below.


\subsection{Articulated Motion}
Articulated motion consists of rigid segments connected by joints that move relative to each other, producing globally non-rigid behavior while each segment undergoes only rigid transformations (translation and rotation). This motion is commonly seen in biological systems (e.g., skeletons) and engineered mechanisms.

Articulated objects are modeled as kinematic trees \cite{abbatematteo2019learning}-hierarchical structures where joints define degrees of freedom (revolute, prismatic, or fixed) and govern transformation propagation through the hierarchy. Methods for articulated motion fall into two categories. 
\textbf{Template-based methods} utilize predefined parametric models that encode a known shape and/or skeletal structure for specific object categories. \textbf{Template-free methods} infer geometry and kinematics directly from observations without category-specific priors, enabling generalization to arbitrary articulated objects at the cost of increased computational complexity and ambiguity in joint inference.

\subsubsection{\gmark{Template-based Articulation}}
\label{subsec:template_articulation}
Among available template choices (see ~\Cref{subsec: template-based representation}), we take SMPL~\cite{SMPL:2015} as an exemplar for the following discussion, as most template-based methods share similar motion modeling mechanisms.
To articulate a standard template mesh from a rest pose to a target pose,
SMPL deforms its template mesh using \textit{linear blend skinning (LBS)}, which assigns blend weights to vertices to quantify each joint's influence. The deformed position $\mathbf{p}_i^m$ of a canonical point $\mathbf{p}_i^c$ is computed as:
\begin{equation}
    \mathbf{p}_i^m = \mathcal{D}_{\theta}(\mathbf{p}_i^c, \mathbf{w}(\mathbf{p}_i^c); \tau(J)) = \sum_{j=1}^{J} w_j(\mathbf{p}_i^c) \mathbf{T}_j \mathbf{p}_i^c,
    \label{eq:lbs_paraphrased}
\end{equation}
where $w_j(\mathbf{p}_i^c)$ is the blend weight for the $j$-th joint and $\mathbf{T}_j \in \mathbb{R}^{4\times 4}$ is its transformation matrix. \mingz{This provides natural support for in-domain applications such as \gmark{human motion synthesis~\cite{PhysDiff} and reconstruction~\cite{goel2023humans}}, where tasks are accomplished through template parameter prediction. Further combination with visually rich representations such as Gaussian Splatting~\cite{li2024animatable} allows realistic motion and appearance to be generated simultaneously.}



\noindent\textit{Neural skinning fields} address off-surface deformation challenges by learning skinning functions directly from data. A neural network maps 3D points to blend weights:
\begin{equation}
    \mathbf{w}(\mathbf{x}) = \mathcal{S}_\theta(\mathbf{p}; \tau(t)),
\end{equation}
where $\tau(t)$  provides time-varying conditions such as pose parameters. \mingz{\gmark{Instead of using the fully equipped SMPL template, this branch of work, at minimum, takes only the skeleton as input and derives the skinning weight. HumanNeRF~\cite{weng2022humannerf}, TAVA~\cite{li2022tava}, and A-NeRF~\cite{peng2021animatable} employ off-the-shelf modules to predict posed skeletons as the prior and model motion sequence from input videos. LS-Avatar~\cite{song2025locality} applies part-based neural skinning fields, enabling realistic avatar animation from monocular video. NeuroSkinning~\cite{liu2019neuroskinning} and SNARF~\cite{chen2021snarf} take the skeletons as input and infer skinning weights to synthesize motion, with the former demonstrating applicability on cloth animations once the skeleton is provided.}}



\subsubsection{Template-Free articulation}
Template-free methods reconstruct articulated objects without predefined models, discovering geometry and kinematics from observations. This presents three challenges: \textbf{(1)} inferring movable parts, \textbf{(2)} identifying joint locations and axes, and \textbf{(3)} estimating motions from limited data—compounded by diverse articulation patterns, occlusions, and sparse observations. Solutions typically employ two approaches: explicit joint parameter estimation or kinematic tree discovery.

\noindent\textbf{Joint Parameters:} Joint parameters provide a structured representation specifying motion properties including joint type, axis, state, and limits \cite{liu2022akb}. Common 1-DoF joints (revolute, prismatic) dominate everyday objects, while multi-DoF joints (e.g., ball joints) appear in complex mechanisms. The joint axis encodes orientation and pivot location; joint state represents current configuration (rotation angle or translation distance) \cite{chignoli2024urdf+}. 

\noindent\textbf{Kinematic Tree:} The kinematic tree provides a hierarchical abstraction capturing part dependencies through explicit structural relationships \cite{lei2023nap, eisner2022flowbot3d}. This graph-based representation—with parts as nodes and joints as edges—proves essential for multi-component objects \cite{chen2024urdformer}. However, topology extraction is challenging due to object-specific variations in part count and structural configuration \cite{le2024articulate, kim2025screwsplat}. Recent advances address this through graph neural networks that learn connectivity patterns and enhanced representations supporting complex structures like kinematic loops \cite{chignoli2024urdf+}.

\subsubsection{Integrating physical priors}
Physical priors ensure plausible reconstructions across both paradigms. Template-based models embed constraints in learned spaces~\cite{SMPL:2015}, model continuous velocity~\cite{siyao2025half}, or learn efficient dynamics~\cite{andriluka2024learned}. Collision detection employs spatial partitioning and CCD~\cite{ccd}.
Template-free methods infer consistency from observations, through enforcing SDF and kinematic constraints~\cite{wu2025reartgs} or incorporating probabilistic limits~\cite{Sturm_2011}. Alternatively, integration via neural priors~\cite{andriluka2024learned} or physics constraint injection~\cite{PhysDiff} enables physically consistent reconstructions for robotics, animation, and simulation.

\subsection{Deformation-guided Representation}

This paradigm factorizes dynamic scenes into a static \emph{canonical space} and time-varying \emph{deformation fields}~\cite{huang2024sc,park2021nerfies,guoNeuralDeformableVoxelGridFast2022a,liu2022devrf,zhao2023animate124,ren2023dreamgaussian4d}. The canonical space serves as a reference from which all frames derive via learned deformations. For rigid or articulated cases, it corresponds to a neutral pose; for general scenes, it preserves sufficient structure for reliable temporal correspondence.

Deformation fields, implemented as neural networks, formalize the canonical-to-observation relationship. Forward deformation $\Phi_\theta$ maps canonical to observation space at time $t$:
\begin{equation}
    \Delta_{b \to f}(\mathbf{p}_b) = \Phi_\theta\bigl(\mathbf{p}_f;\ \tau(t)\bigr),
\end{equation}
while backward deformation $\Phi_\theta^{-1}$ provides the inverse \snc{and $\tau(t)$ is a function of time.} Invertibility ensures bidirectional consistency through regularization or invertible architectures~\cite{liu2025modgs,wangTrackingEverythingEverywhereAllOnce2023,caiNeuralSurfaceReconstructionDynamicScenes2022}. NeRF frameworks typically use backward deformation to query canonical attributes~\cite{park2021nerfies,pumarola2021d,park2021hypernerf}, while explicit representations like 3D Gaussians employ forward warping~\cite{yang2024deformable,liu2025modgs}. For long sequences with substantial motion, multiple local canonical spaces over temporal subwindows maintain coherence while handling large transformations~\cite{attal2023hyperreel,nag20252}.

Deformation-based motion representations in 4D tasks encounter key limitations. 
\textit{Topological variations} such as object emergence, disappearance, or merging break the bijective mapping assumption, leading to correspondence failures~\cite{li2021neural,tretschk2021non}. 
\textit{Large inter-frame motions} challenge the smoothness and invertibility of learned deformations, often entangling geometry and appearance~\cite{park2021hypernerf,pumarola2021d}. 
\textit{Ambiguous correspondences} arise under occlusion or low-texture regions where multiple deformations fit equally well~\cite{yuan2021star,wang2023neuralangelo} while \textit{computational overheads} grow rapidly with sequence length and bidirectional consistency~\cite{yang2024deformable,luiten2024dynamic}. 
Lastly, \textit{non-rigid or fast motions} remain difficult to capture due to temporal discretization limits~\cite{peng2021neural,li2022neural}.
Deformation fields model two main types of motion: (a) rigid and (b) non-rigid motion. 

\subsubsection{Rigid Motion}

Rigid motion preserves geometry through rotation and translation, maintaining shape, size, and internal distances. A rigid transformation is expressed by a rotation matrix $\mathbf{R} \in SO(3)$ (or quaternion $\mathbf{q} \in \mathbb{R}^{4}$) and translation vector $\mathbf{t} \in \mathbb{R}^3$:
\begin{equation}
    \mathbf{p}_{t} = \mathcal{D}_{\theta}\bigl(\mathbf{p}_{t-1}\bigr) = \mathbf{R}\,\mathbf{p}_{t-1} + \mathbf{t}.
\end{equation}
Rigid bodies include manufactured items (boxes, chairs, kitchenware) and vehicles, which are often approximated as rigid despite minor internal deformations~\cite{peng2022cagenerf,yifan2020neural,xu2022deforming}. 

\subsubsection{Non-Rigid Motion}
Non-rigid motion involves local point deformations essential for modeling cloth~\cite{qiu2023rec}, facial expressions~\cite{wang20253d}, and fluids~\cite{gao2025fluidnexus}. Unlike rigid or articulated motion with structured parametric models~\cite{SMPL:2015}, generic non-rigid deformation lacks compact formulations due to topological changes and infinite degrees of freedom.

Recent approaches model non-rigid motion as learnable neural fields estimating continuous displacement:
\begin{equation}
    \mathbf{p}_t = \mathcal{D}_{\theta}\bigl(\mathbf{p}_{t-1};\tau(t)\bigr)
    = \mathbf{p}_{t-1} + \Delta_{\theta}\!\bigl(\mathbf{p}_{t-1};\tau(t)\bigr),
\end{equation}
where \(\Delta_{\theta}\) predicts displacements conditioned on temporal code \(\tau(t)\). These fields are trained from 2D observations via differentiable rendering, eliminating 3D supervision. However, sparse views complicate learning, requiring regularization for robustness. 

\subsubsection{Integrating physical priors}
Physics-based 3D deformation integrates simulation with neural representations. Material Point Method (MPM)~\cite{stomakhin2013material} handles large deformations and topology changes. PhysGaussian~\cite{xie2024physgaussian} treats 3D Gaussians as simulation particles, which has been extended to language-driven property assignment~\cite{qiu2024feature} and text-to-3D synthesis. NeRF methods embed continuum mechanics: PIE-NeRF~\cite{feng2024pie} and PAC-NeRF~\cite{li2023pac} enable interactive elastodynamics, while Nerfies~\cite{park2021nerfies} regularizes deformation with elastic priors. Alternative frameworks explore finite elements, spring-mass systems~\cite{zhong2024reconstruction}, and position-based dynamics~\cite{abou2023physically}. ~\cite{chen2025physics} uses elasticity as a constraint in point cloud modelling, while differentiable physics~\cite{huang2024dreamphysics,li2023nvfi} leverages neural networks to distill physical laws.

\subsection{Tracking-based Representation}

Rather than relating observations to a shared reference, tracking-based methods capture motion between consecutive frames. This frame-to-frame approach leverages incremental deformations, naturally accommodating topology changes and extreme deformations that challenge canonical-space methods.

\subsubsection{2D Tracking}
Reliable 2D tracking establishes temporal correspondences foundational for 3D motion recovery. Classical methods include sparse feature matching (e.g., ORB~\cite{rublee2011orb}) and optical flow~\cite{sun2018pwc,teed2020raft}. Sparse matching supports localization but fails for dense non-rigid motion; optical flow provides dense correspondences but struggles with occlusion and large viewpoint changes. Recent neural rendering methods~\cite{wang2021neural,wangTrackingEverythingEverywhereAllOnce2023,luiten2024dynamic} enable dense, long-range 3D tracking by jointly optimizing scene representation and a continuous trajectory field:
\begin{equation}
    \mathbf{p}_t = \mathcal{J}_\theta(\mathbf{p}_{t-1}; \tau(t)),
\end{equation}
Here, $\mathcal{J}_\theta$ models per-point 3D motion guided by 2D tracking priors $\tau(t)$, achieving global temporal coherence~\cite{chen2023periodic,stearns2024dynamic,wangshapemotion4dreconstructionsingle2024c}.

Despite its accessibility, 2D tracking faces several critical challenges for 4D tasks:
\textit{(a) Depth ambiguity:} 2D projections lose 3D geometric information, making it difficult to distinguish between different 3D motions that produce similar 2D displacement.
\textit{(b) Occlusion handling:} Points disappear and reappear in 2D views, breaking temporal correspondence.
\textit{(c) Perspective distortions:} 2D tracking doesn't capture actual 3D motion—camera viewpoint changes confound object motion.
\textit{(d) Requires lifting:} An additional step is needed to reconstruct 3D/4D information from 2D observations, introducing errors and ambiguity.

3D tracking and scene flow overcome these limitations by directly estimating motion in 3D space, preserving geometric information throughout the temporal sequence.

\subsubsection{3D Tracking} 
The 3D \textit{scene flow} provides a compact representation of frame-to-frame motion in 4D reconstruction. 
In the continuous setting, a velocity field $\mathbf{v}(\mathbf{x}, t)$ assigns each point a motion vector describing its instantaneous change~\cite{niemeyer2019occupancy,wang2023flow}, where a point’s new position can be approximated as 
\begin{equation}
    \mathbf{p}_t 
    \;=\; 
    \mathbf{o}_{t-1}
    \;+\;
    \int_{t-1}^{t} \mathbf{v}\bigl(\mathbf{p}(\hat{t}), \hat{t}\bigr)\, d\hat{t},
\end{equation}
where $\mathbf{v}(\mathbf{x}(\hat{t}), \hat{t})$ denotes the velocity at intermediate time $\hat{t}$. Although theoretically elegant, acquiring continuous ground-truth velocities is typically infeasible in practice. 
In discrete form, the scene flow $\mathbf{O}(\mathbf{p}, t) = \mathbf{p}_t - \mathbf{p}_{t-1}$ directly encodes per-point displacements between frames, serving as the practical counterpart of continuous velocity fields for learning temporally coherent 4D motion.

Scene flow offers key \textit{advantages}: it resolves depth ambiguity in 2D tracking, ensures geometrically consistent 3D motion, and naturally handles topology changes~\cite{niemeyer2019occupancy}. 
Neural methods such as NeRFlow~\cite{du2021neural} couple flow and radiance fields for continuous spatiotemporal modeling, while Neural Scene Flow Fields~\cite{li2021neural} predict forward and backward flows to capture sharp motion boundaries and enable smooth interpolation.

However, scene flow methods have notable \textit{limitations}: they are sensitive to occlusions and visibility constraints, leading to tracking failures~\cite{lin2022occlusionfusion}. 
Highly deformable or low-texture scenes exacerbate self-occlusion and reduce accuracy~\cite{duisterhof2023deformgs}, while frame-wise processing struggles with topology changes and temporal inconsistency. 
Moreover, limited sensor visibility restricts motion estimation to observed surfaces, leaving discontinuities in the reconstructed motion field~\cite{li20214dcomplete}.

\subsubsection{Integrating physical priors}
Integrating physics into 2D tracking regularizes the ill-posed monocular reconstruction problem through \textit{motion constraints and physical priors}. \textit{Optical-flow-based methods}~\cite{wangshapemotion4dreconstructionsingle2024c,gao2024gaussianflow} combine flow and depth cues with physical motion models to enforce coherent spatio-temporal dynamics. \textit{Deformation-based approaches} apply physics-inspired constraints including as-rigid-as-possible (ARAP) regularization~\cite{lei2024mosca} to preserve local rigidity. Meanwhile, \textit{learning-based frameworks}~\cite{feng2025st4rtrack,niemeyer2019occupancy} exploit 2D correspondences and monocular depth through reprojection or implicit dynamic modeling, achieving 4D reconstruction without explicit supervision.

Physics integration in 3D tracking depends on the \textit{choice of 3D representation}. For \textit{rigid-body/mesh models}, physics-based filters like~\cite{Kandukuri_2024} embed differentiable simulators into Extended Kalman Filters (EKF) to track pose, velocity, and friction, while~\cite{piga2021differentiable} applies differentiable EKFs to simpler 1D sliding motion. \textit{Gaussian/particle representations} couple dynamics with rendering:~\cite{luiten2024dynamic} models scenes as moving 3D Gaussians with local rigidity constraints,~\cite{zhang2024dynamic} learns dynamics via Graph Neural Networks on sparse particles, and~\cite{xie2024physgaussian} enriches Gaussians with Newtonian mechanics (strain, stress, inertia). For \textit{extended object tracking},~\cite{Kumru_2021} uses Gaussian processes to jointly infer shape and kinematics, while~\cite{Barad_2024} employs object-centric Gaussian splatting to reconstruct and track dynamic objects from RGBD.

\subsection{Hybrid Representation}
\saura{While we have examined explicit motion representations that integrate motion into geometric structures, several approaches extend beyond this paradigm—either by leveraging multiple concurrent motion models or by adopting alternative formulations for motion representation. We discuss them below.}

\noindent \textbf{Joint representation} Most real-world environments involve a blend of motion types—rigid, articulated, and non-rigid—whose coexistence gives rise to hybrid dynamics. The human body is a canonical example: an articulated skeleton undergoes global rigid motion, while soft tissue, loose garments, and hair introduce highly non-rigid variations~\cite{qiu2023rec,jiang2022selfrecon}. Modeling such interplay requires a representation that enforces global structure yet retains flexibility for localized deviations.

A common strategy is to decompose motion into complementary components:
\begin{equation}
    \mathbf{p}_t
    \;=\;
    \underbrace{\mathcal{T}_{\theta_1}\bigl(\mathbf{p}_{t-1};\tau(t)\bigr)}_{\text{coarse, e.g.\ rigid/articulated}}
    \;+\;
    \underbrace{\Delta_{\theta_2}\bigl(\mathbf{p}_{t-1};\tau(t)\bigr)}_{\text{fine, e.g.\ non-rigid residual}}.
\end{equation}
Here, $\mathcal{T}_{\theta_1}$ denotes a structured, interpretable global motion model (e.g., rigid or articulated transformations), while $\Delta_{\theta_2}$ is typically realized as a neural field that predicts residual displacements for local deformations. Beyond human and animal motion, this paradigm has been extended to deformable objects with near-rigid parts and to multi-object scenes exhibiting heterogeneous dynamics~\cite{chenOmniReOmniUrbanSceneReconstruction2024a,fischer2024dynamic}. Such hierarchical factorization offers several benefits: it preserves interpretability by isolating coarse transformations; simplifies learning by restricting the neural component to residuals; and enables explicit control over global motion while retaining the capacity to model fine-scale variations.

\noindent\textbf{4D space-time} 
4D space–time methods represent dynamic scenes as unified spacetime volumes using explicit 4D primitives \cite{duan20244d,yang20244d}, where motion is inherently encoded within the 3D representation itself, eliminating the need for deformation fields \cite{katsumata2024compact}. Geometry, appearance, and motion are jointly modeled in a single neural field, allowing density, color, and opacity to evolve over time without explicit motion vectors. Formally,
\begin{equation}
    I_o 
    = 
    \mathcal{R}_{\theta}\bigl(\mathbf{p}, \tau(g,t)\bigr)
\end{equation}
where $\mathcal{R}_{\theta}$ is the rendering function, $\tau(g,t)$ captures geometric variations such as 4D $\sum,\sigma$ in Gaussian Splatting or color density in NeRFs, and $t$ denotes direct time input (e.g., frame index or learnable per-frame latents). The key design choice lies in modeling $\tau(g,t)$: feed-forward models \cite{xu20254dgt,ma20254d} decode per-pixel 4D Gaussian vectors via DiT heads, while \cite{duan20244d,yang20244d} instead use 1D temporal Gaussians. NeRF-based approaches progress from MLPs to 4D neural voxels for faster rendering \cite{park2023temporal,gan2023v4d}. In all cases, motion is implicitly embedded in the scene’s geometry and supervised purely from 2D images, making this paradigm computationally efficient and well-suited to recent feed-forward 4D generation models \cite{ma20254d,xu20254dgt}. 

\noindent\textbf{Per-frame modeling} Per-frame 4D reconstruction methods treat each timestep as an independent or weakly coupled 3D reconstruction task \cite{cut3r}, producing discrete per-frame 3D representations that are subsequently assembled into a 4D sequence. L4GM \cite{ren2024l4gm} generates frame-wise Gaussian representations from LGM \cite{tang2024lgm} conditioned on predictions from previous timesteps, while point-based approaches \cite{cut3r,chen2025easi3r,zhang2024monst3r} adopt DusT3R \cite{wang2024dust3r} independently for each frame to obtain 4D outputs. However, this frame-by-frame generation strategy typically necessitates interpolation or post-processing to enforce temporal smoothness. For example, \cite{ren2024l4gm} employs a video interpolation model \cite{blattmann2023align} to refine renderings, and \cite{chen2025easi3r} incorporates temporal attention modulation to enhance overall reconstruction fidelity.


\section{Modeling Interaction}
\label{sec:interac}

\begin{figure*}
    \includegraphics[width=\linewidth]{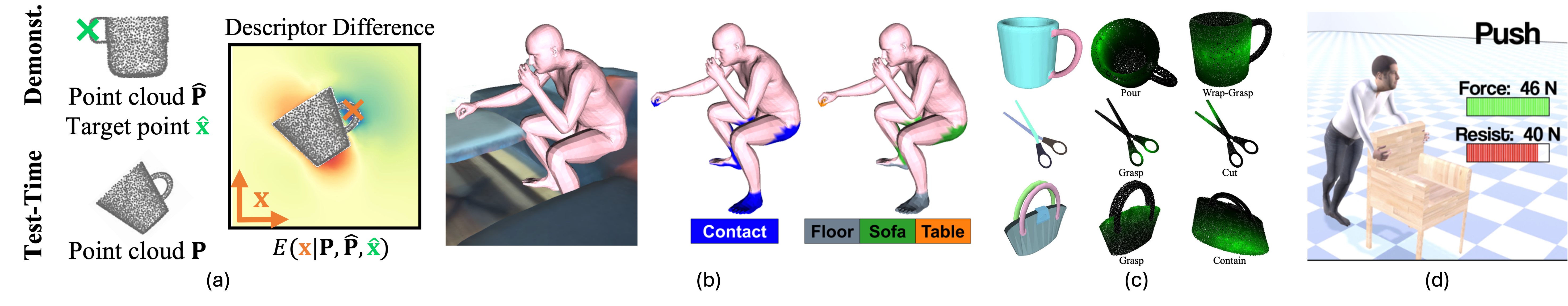}
    \caption{Interaction specific representations. (a) Parametrizing poses between interaction entities~\cite{NeuralDescriptorFields}; (b) Representing contacts~\cite{POSA}; (c) Representing affordances~\cite{3DAffordanceNet}; (d) Representing physical properties~\cite{FORCEIntuitivePhysics}.}
    \label{fig: interaction}
\end{figure*}

Thus far, we have focused on modeling the geometry and motion of 4D entities, in isolation.
It is often necessary, however, to model the interactions between multiple 4D entities: between human and human, human and object, or object and object. 
\mz{Such interactions can be learned either from data priors~\cite{lu2025humoto} where knowledge of dynamics and causality is implicitly captured from large-scale observations, or from physical simulations~\cite{li2020codimensional} where interactions are explicitly governed by forces, materials, and physics-based rules. \gmark{We refer readers to surveys dedicated to these two perspectives~\cite{zhu2023MotionGeneration, sui2026survey, fan20253DInteractionSurvey}.}
In this section, we take a complementary angle, surveying explicit interaction representations that describe the structure of interactions, independent of specific objects or scenes (\Cref{fig: interaction}).}
We begin by surveying the overall representation strategies for entities involved in interactions.
We then delve deeper into three aspects crucial in interaction: pose, contact, action and affordance.
Finally, we examine strategies for integrating physical priors into interaction representations.

\subsection{Representing Interaction Entities}
Interaction occurs between multiple entities, each having its individual motion.
Naturally, one could model an interaction $I$ involving $N$ entities as a collection of motion sequences
\[
I = \{ M_1, M_2, \ldots, M_N \},
\]
where each $M_i = \{ \mathbf{x}_i^1, \mathbf{x}_i^2, \ldots, \mathbf{x}_i^T \}$ 
is a sequence of states $x_i^t$ describing the motion of entity $E_i$ over time steps $1 \ldots T$.
The representation choice of each entity depends on both its individual properties, as well as the property of the entire interaction. 
Humans are most commonly represented with skeletal poses or template meshes (Section~\ref{subsec: template-based representation}), as these excel at modeling long motion sequences which are central to most interaction events. 
Given the complexity of interaction data, simpler representations are sometimes favored to make the problem tractable.
To address the weaknesses of template meshes, 
a growing trend is to couple them with neural field based representations to obtain increased geometric fidelity~\cite{pokhariya2024manus, qu2025hogsa} and more flexible, differentiable optimization of poses and contact objectives~\cite{on2025bigs, wen2025reconstructing, li2024zerohsi, cao_avatargo_2024}.
For other objects, meshes (Section~\ref{subsec: mesh rep}) are the popular choice due to having explicit surface representations, which enables more efficient modeling of contact and physics.
\mz{Despite these advances, the assumption of object rigidity remains a major simplification. Extending to articulated objects brings more flexibility, but also introduces the need for a unified representation that captures part connectivity, motion patterns, and pose variations beyond fixed templates.}

\subsection{Pose}
To accurately model interactions, it is crucial to precisely describe the poses of the interacting entities.
The fundamental design choice is between storing poses in global coordinates, or storing relative poses.
While single-entity methods often canonicalize poses by removing global position and orientation, recent interaction-focused works preserve global coordinates to maintain spatial relationships~\cite{InterX, bhatnagar22behave, Intergen}.
This preservation is essential because relative positions and orientations between entities \emph{are} the interaction signal---canonicalizing to a single reference frame discards this critical information.
Independent of coordinate choice, the rotation parametrization itself requires careful consideration.
Most modern works adopt 6D continuous rotation~\cite{YiZhouRotation} instead of traditional representations (quaternions, axis-angle, Euler angles) to avoid discontinuities during neural network optimization.
While global coordinates preserve spatial context, they create challenges: similar interactions at different world locations have completely different coordinate values.
The solution is to store global poses but compute relative features for learning.
Common relative features include inter-entity distances, relative orientations, and facing directions.
Beyond these handcrafted features, learned representations like Neural Descriptor Fields~\cite{NeuralDescriptorFields} provide SE(3)-equivariant encodings that generalize across different global configurations, while maintaining relative spatial relationships.
Hybrid representations~\cite{GOALGrasping} that combine global poses and relative features is an interesting direction in addressing the challenges for pose representation.

\subsection{Contact}
Interacting entities make contact with each other.
Accurately representing contact can either be an end goal, or a means for more accurate interaction modeling.
Explicit representations of contact, such as \textit{contact maps}~\cite{ContactDB, ContactPose, GRAB:2020} and \textit{neural distance fields}, have been shown to improve motion generation. 
\cite{mao2022contact} shows that predicting future contact maps enhances both local and global motion consistency. 
GanHand~\cite{corona2020ganhand} refines contact modeling at finger and force levels to improve pose estimation. 
Neural distance fields (NDF) emerges as an auxiliary contact representation. 
NDF~\cite{karunratanakul2020grasping,tiwari2022pose,weng2022neural,urain2022se,kulkarni2024nifty} expands the single-frame local contact information into a spatially-dense field to guide motion generation. 
Grasping Fields~\cite{karunratanakul2020grasping} encodes the distances to valid grasps, and PoseNDF~\cite{tiwari2022pose} learns to encode full-body pose into the field. 
NIFTY~\cite{kulkarni2024nifty} encodes the body pose gradient towards interacting frame into the fields. 
All these works demonstrate that explicit contact modeling improves generation.
Most existing research, however, relies on SMPL or MANO, which are rigid representations and naturally induce penetration. 
Another future step might be to extend contact modeling beyond the visual domain to incorporate mechanical realism and force reasoning.

\subsection{Action and Affordance}
While geometry, poses and contact implicitly describe the type of actions, it is also common to explicitly model the affordances~\cite{Gibson} and actions, describing both possibility of interaction and the actual interaction events.
Such actions and affordances are commonly represented with explicit~\cite{PiGraphs} or neural implicit~\cite{NeuralStateMachine} graph structures.
It is also common to directly add additional labels (e.g. "hook pull", "key press") either at object/part level~\cite{SceneFun3D} or as dense per point labels~\cite{3DAffordanceNet, PartAffordance}.
Motion/articulation annotations can be added for affordances that involve movable parts.
Instead of explicit labels, one could also adopt a field-based representation~\cite{AffordMotion, kulkarni2024nifty} to support continuous queries.
Other types of queries, such as motion parameters, force, etc, can also be integrated into such representations.
With the emergence of foundation models, more works have opted to adopt open-vocabulary representations~\cite{LASOAffordance, AffordanceLLM}.

\subsection{Ensuring Physical Plausibility}
Physical plausibility is also fundamental for interaction related tasks. It is important to ensure plausible contact without floating and penetration, accurate object dynamics with respect to various forces, and valid interaction with the environment overall.
A wide range of techniques have been adopted to ensure such physical plausibility.
First, one could directly apply physics and interaction aware losses during training and optimization~\cite{corona2020ganhand,jiang2021hand,grady2021contactopt,yang2021cpf,turpin2022grasp, ReGenNet}.
Metrics include foot sliding, average and maximum penetration depth or volume, hand-to-object distance, contact IoU and normal alignment, etc.
It is also possible to perform physics simulation and contact-based optimization, and incorporate them through post-processing~\cite{COINS}, correction steps~\cite{PhysDiff, xu2023interdiff}, learnable surrogates~\cite{DeepSimHo}, or combination with reinforcement learning~\cite{D-Grasp, hassan2023synthesizing}. 
Another line of work directly injects physics-aware encoding into the network structure, encoding physical properties like force, resistance and contact~\cite{FORCEIntuitivePhysics}.

\section{Datasets and Benchmarks}
\label{sec: dataset}

\begin{figure*}
    \centering
    \includegraphics[width=\linewidth]{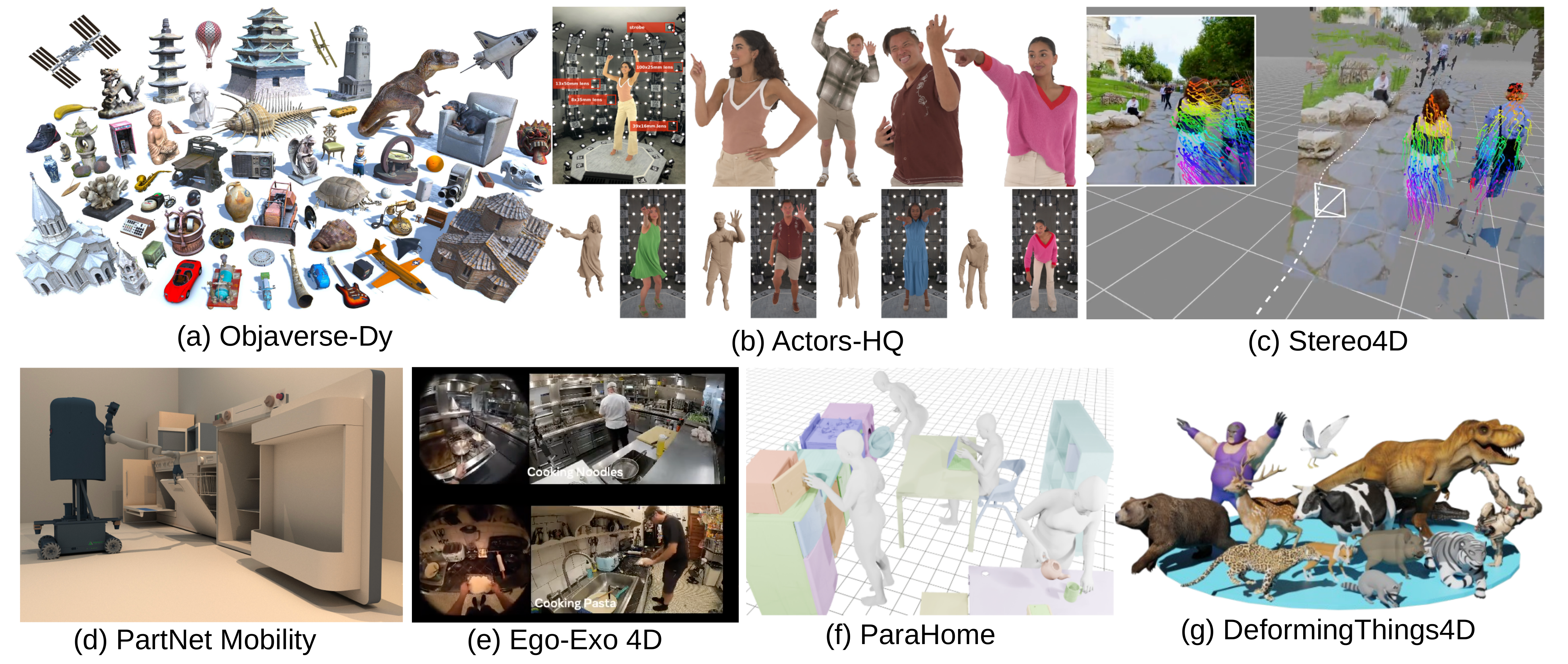}
    \caption{Overview of key datasets in 4D representation research: (a) Objaverse-Dy~\cite{yao2025sv4d} provides a large-scale collection of synthetic animated 3D assets, serving as the primary source for current 4D generation works after filtering; (b) Actors-HQ~\cite{icsik2023humanrf} is a high-fidelity, multi-view human performance capture dataset with per-frame 3D mesh reconstructions for dynamic human motion; (c) Stereo4D~\cite{jin2024stereo4d} represents emerging large-scale datasets capturing native 4D data from real-world scenarios; (d) PartNet-Mobility~\cite{Xiang_2020_SAPIEN} offers an articulated object database that facilitates part-based representation and articulated motion learning; (e) Ego-Exo4D~\cite{grauman2024ego} is a multi-camera synchronized video corpus capturing first-person (egocentric) and third-person (exocentric) views of skilled human activities; (f) Parahome~\cite{kim2025parahome} is a dataset capturing 3D human–object interaction in a natural home environment, and (g) DeformingThings4D~\cite{li20214dcomplete} is a large-scale synthetic dataset of non-rigidly deforming 3D objects (humanoids and animals) with dense 4D annotations.}
    \label{fig:dataset}
    \vspace{-0.5em}
\end{figure*}

Unlike other computer vision tasks with abundant labeled data, 4D representation learning faces a data scarcity problem. Ideal datasets would contain real-world dynamic scenes with complete ground-truth geometry, appearance, motion and interaction annotations. However, such datasets remain limited due to technical and economic challenges in capturing high-quality spatio-temporal data. To address this, the current 4D field leverages a mixture of datasets ranging from 2D to 4D dimensions to guide 4D representation learning ~(\Cref{fig:dataset}). This section surveys the current dataset in use from the perspective of learning geometry, motion and interactions.

\subsection{Geometry Datasets}
\label{subsec:geometry_datasets}

Geometric learning in 4D works primarily leverages two data paradigms: static 3D object datasets for feed-forward reconstruction and large-scale 2D image collections for score-distillation-based generation. 

\noindent\textbf{Static 3D Object Datasets.} Modern 4D generation systems~\cite{chen2025v2m4,song2025puppeteer} employ feed-forward 3D generation models~\cite{xiang2025structured,li2025triposg,zhao2025hunyuan3d} to first synthesize static geometry, then animate it through video guidance or rigging techniques. These feed-forward models are enabled by large-scale 3D object collections. Primary datasets in early time include ShapeNet~\cite{chang2015shapenet} (51K objects) and ModelNet~\cite{wu20153d} (12K objects), while the more recent Objaverse-XL~\cite{deitke2024objaverse} dramatically scales to 10+ million models, becoming one of the most widely adopted datasets in current research. However, Objaverse exhibits significant quality variance, leading researchers to create high-quality filtered subsets~\cite{xiang2025structured,zhang2024clay}. Other smaller high-quality 3D collections, including ABO~\cite{collins2022abo}, 3D-Future~\cite{fu20213d}, OmniObject3D~\cite{wu2023omniobject3d}, Toy4K~\cite{stojanov2021using}, HSSD~\cite{khanna2024habitat}, and GSO~\cite{downs2022google} are commonly used as training set additives or benchmark datasets. These datasets provide native supervision for high-quality 3D reconstruction.

\noindent\textbf{Large-Scale 2D Image Collections.} Score distillation sampling (SDS) based 4D generation~\cite{bahmani20244dfy,ling2024align,singer2023text} leverages foundational image diffusion models to generate 2D supervision for 3D scene reconstruction through volumetric rendering. The generalizability of this paradigm—enabling arbitrary 4D scene generation from text prompts—is inherited from web-scale training on LAION-5B~\cite{schuhmann2022laion} (5.85 billion image-text pairs). However, standard image diffusion models lack 3D awareness, causing multi-face artifacts (the Janus problem~\cite{armandpour2023re}). To address this, multi-view diffusion models~\cite{shi2023mvdream,liu2023zero,shi2023zero123++,liu2023syncdreamer} are fine-tuned using rendered multi-view images from filtered Objaverse objects, significantly improving geometric reconstruction quality in 4D pipelines through enhanced multi-view consistency.

\subsection{Motion Datasets}
\label{subsec:motion_datasets}

Temporal dynamics learning in 4D generation relies on four primary data sources: large-scale video collections for learning general motion patterns, specialized multi-view captures for high-fidelity scene dynamics, parametric motion libraries that provide explicit temporal supervision and deformable object collections for native non-rigid motion learning.

\noindent\textbf{Large-Scale Video Collections.} Video diffusion models serve as key components for guiding motion synthesis in 4D works~\cite{li2024vivid,li2024puppet,vora2025articulate,jiang2024animate3d}. Unlike images that capture only static geometry, videos encode both geometric structure and temporal dynamics. WebVid-10M~\cite{bain2021frozen} and HD-VILA-100M~\cite{xue2022advancing} represent two prominent large-scale video datasets that enable training models like Align Your Latents~\cite{blattmann2023align} and Make-a-Video~\cite{singer2023makeavideo}. These foundational video diffusion models have been successfully adapted for 4D generation~\cite{singer2023text,ling2024align}. Following strategies used to enhance image diffusion models, recent work finetunes video diffusion models~\cite{xie_sv4d_2024,yao2025sv4d,wu2025animateanymesh,liang_diffusion4d_2024} on filtered dynamic objects from Objaverse, significantly improving multiview motion consistency. The primary advantage of video diffusion models lies in their generalizability from large-scale training data, enabling motion synthesis across diverse applications. However, 2D video supervision prioritizes frame-level visual realism over 3D geometric constraints, often producing sequences that appear locally plausible but violate physical continuity when reconstructed as coherent 3D trajectories.

\noindent\textbf{Specialized Multi-View Captures.} Domain-specific datasets provide high-fidelity temporal ground truth through controlled multi-view recording systems. For human motion, representative datasets include ZJU-MoCap~\cite{peng2021neural}, DynaCap~\cite{habermann2021real}, ActorsHQ~\cite{icsik2023humanrf}, and Human3.6M~\cite{ionescu2013human3}. For general dynamic scenes, D-NeRF~\cite{pumarola2021d}, Nerfies~\cite{park2021nerfies}, HyperNerf~\cite{park2021hypernerf}, and Plenoptic Video~\cite{li2022neural} capture real-world deformations with varying complexity. These datasets preserve authentic physical dynamics and provide strong multi-view supervision for consistent 3D motion reconstruction. However, limited scale—typically dozens to hundreds of sequences compared to millions in large-scale video collections—and substantial capture infrastructure requirements constrain their broader applicability.

\noindent\textbf{Animated Motion Libraries.} Synthetic motion captures offer perfect ground truth with structural priors for controllable animation. AMASS~\cite{mahmood2019amass} consolidates over 40 hours of motion capture from 15 datasets into unified SMPL~\cite{SMPL:2015} parametric representations for human motion synthesis. Mixamo~\cite{mixamo} provides thousands of professionally rigged character animations, while Truebones Zoo~\cite{Truebones} extends coverage to 75+ non-humanoid creatures. For animated objects, curated Objaverse subsets~\cite{liang_diffusion4d_2024,xie_sv4d_2024,li2024puppet,li2024vivid} provide hundreds of thousands of animated assets filtered by motion quality and temporal coherence. Articulation-XL~\cite{xiang2025structured} provides 48K+ models featuring automated skeleton quality assessment via Vision-Language Models. These synthetic collections enable large-scale training with explicit motion annotations, though they exhibit domain gaps and simplified physics compared to real-world capture, particularly for complex interactions and environmental effects.

\noindent\textbf{Deformable Object Collections} provide ground truth for non-rigid motion learning. DeformingThings4D~\cite{li20214dcomplete} contains 1,972 synthetic animation sequences spanning 31 categories (humanoids and animals) with dense 4D annotations including signed distance fields, volumetric motion fields, and scene flow at multiple hierarchical resolutions. The most recent effort, Stereo4D~\cite{jin2024stereo4d} offers 100K+ real-world clips mined from Internet VR180 videos, providing pseudo-metric 3D point clouds with long-range world-space trajectories derived through careful fusion of stereo depth estimation, 2D tracking, and structure-from-motion. These datasets address the scarcity of real-world 4D supervision—DeformingThings4D through high-quality synthetic rendering and Stereo4D through scalable data collection.

\subsection{Interaction Dataset}
\label{subsec: interaction}
Interaction-focused datasets capture relationships between objects, humans, and environments through three approaches: articulated object collections that encode structural priors, human-object interaction datasets capturing co-evolving dynamics, and multi-entity scenarios modeling complex relational behaviors.

\noindent\textbf{Articulated Object Collections.} Structured representations of articulated objects provide functional-aware supervision for controllable 4D generation. Available datasets span multiple sources: synthetic object-centric collections include PartNet-Mobility~\cite{Xiang_2020_SAPIEN} and its successors ACD~\cite{iliash2024s2o} and Phys-X 3D~\cite{cao2025physx}; real-world scans include object-centric AKB-48~\cite{liu2022akb} and scene-level Multi-Scan~\cite{mao2022multiscan} with articulatable components; and GAPartNet~\cite{geng2023gapartnet} combines both synthetic and real sources. These datasets provide kinematic constraints and structural annotations essential for physically plausible 4D animation, though coverage remains limited to specific object categories with explicit articulation mechanisms.

\noindent\textbf{Human-Object Interaction.} \mingz{HOI datasets capture the coupled dynamics between humans and manipulated objects during interaction sequences, spanning multiple levels of granularity. At the finest level, hand-object interaction datasets including DexYCB~\cite{chao2021dexycb}, ARCTIC~\cite{fan2023arctic}, OAKINK2~\cite{zhan2024oakink2}, and TACO~\cite{liu2024taco} capture single-hand or bi-manual grasping and manipulation with rigid objects, while Obman~\cite{hasson2019learning} provides synthetic data pairing MANO~\cite{MANO:SIGGRAPHASIA:2017} hand models with ShapeNet~\cite{chang2015shapenet} objects. Whole-body interaction datasets extend this scope, with CIRCLE~\cite{araujo2023circle} capturing reaching motions in virtual home environments, COUCH~\cite{zhang2022couch} and CHAIRS~\cite{jiang2023full} focusing on sitting-based furniture interactions, and GRAB~\cite{GRAB:2020}, BEHAVE~\cite{bhatnagar22behave}, and OMOMO~\cite{li2023object} providing diverse activities such as grasping, carrying, and moving household objects. At the broadest level, Human-Scene Interaction (HSI) datasets such as ParaHome~\cite{kim2025parahome}, SAMP~\cite{hassan2021stochastic}, TRUMANS~\cite{jiang2024scaling}, and HUMANISE~\cite{wang2022humanise} incorporate environmental navigation by capturing humans moving through spaces while interacting with multiple scene objects. These datasets enable learning of contact-rich interactions and object affordances, though capture complexity and annotation requirements limit dataset scale and diversity. Given the breadth of this field, the datasets presented here are not exhaustive; we refer readers to~\cite{zhu2023MotionGeneration} for a comprehensive discussion of HOI datasets.}

\noindent\textbf{Multi-Entity Scenarios.} Large-scale datasets capturing multiple interacting entities provide supervision for complex relational dynamics. Egocentric datasets like Ego4D~\cite{grauman2022ego4d} and Ego-Exo4D~\cite{grauman2024ego} capture first-person interaction perspectives during daily activities with synchronized multi-view data. Autonomous driving datasets including Waymo~\cite{sun2020scalability,ettinger2021large} and nuScenes~\cite{caesar2020nuscenes} provide dense spatio-temporal supervision for multi-agent urban scenarios with point clouds and 3D trajectories. Social interaction datasets such as CMU Panoptic~\cite{Joo_2017_TPAMI} and CHI3D~\cite{fieraru2020three} focus on multi-person scenarios. These datasets enable learning of relational behaviors and spatial reasoning, though their domain-specific focus limits direct transfer to general 4D generation tasks.

\subsection{Benchmarks and Evaluation Metrics}
\label{subsec:benchmarks_metrics}

4D evaluation requires assessment across geometric fidelity, temporal consistency, and semantic alignment, progressing from reconstruction-centric evaluation with clear ground truth to generation-centric assessment emphasizing perceptual quality.

\noindent\textbf{Specialized Benchmarks.} Recent benchmarks target specific 4D challenges: WideRange4D~\cite{yang2025widerange4d} focuses on wide-range spatial object movements in synthetic scenes, Inter3D~\cite{chen2025inter3d} benchmarks human-interactable 3D object reconstruction, SEED4D~\cite{kastingschafer2025seed4d} provides annotated driving videos with synchronized ego-exo captures for autonomous driving tasks; InterAct~\cite{xu2025interact} provides 22 hours of HOI data for 6 HOI task benchmarks; 4D-Bench~\cite{zhu20254d} benchmarks MLLM understanding on 4D by assessing captioning abilities. Despite progress, unified benchmarks for core generation tasks (text-to-4D, image-to-4D, dynamic NeRF reconstruction) remain absent. Current practice relies on small-scale datasets like Consistent4D~\cite{jiang2023consistent4d} and D-NeRF~\cite{pumarola2021d} without standardized protocols.

\noindent\textbf{Quantitative Metrics.} Evaluation metrics span multiple objectives. \textit{Appearance fidelity} uses PSNR and SSIM for pixel-level correspondence, complemented by perceptually-aligned LPIPS~\cite{zhang2018perceptual} that better correlates with human judgments. \textit{Temporal consistency} employs Fréchet Video Distance (FVD) on 4D video renders~\cite{jiang2023consistent4d,zeng2024stag4d}, STREAM~\cite{kim2024stream} for disentangled evaluation of temporal coherence, visual fidelity, and diversity without length constraints, and VBench~\cite{huang2024vbench} providing comprehensive assessment across 16 dimensions, including subject consistency and motion smoothness. \textit{Semantic alignment} measures input-output consistency through CLIP-score~\cite{CLIP} for text-to-image tasks and R-Precision for text-to-motion alignment via retrieval accuracy. \textit{Geometric integrity} employs Chamfer Distance for bidirectional point cloud similarity, Earth Mover's Distance (EMD)~\cite{rubner1998metric} for optimal transport cost between point distributions, 3D IoU for volumetric overlap, Hausdorff Distance for maximum deviation, and physics-based constraints for interaction validity~\cite{lu2025humoto}. \textit{Human evaluation} complements automated metrics, assessing faithfulness, aesthetic quality, and physical plausibility through structured user studies~\cite{nag20252,zhang2024physdreamer}.

\noindent\textbf{Current Limitations.} Contemporary 4D evaluation faces fundamental challenges. \textit{Geometric functionality assessment} remains inadequate, with metrics emphasizing shape alignment while overlooking functional correctness, structural balance, and design complexity. \textit{Long-duration motion evaluation} lacks appropriate protocols, as temporal metrics assess only short sequences (<32 frames) without extended temporal consistency evaluation. \textit{Unified benchmark scarcity} limits systematic progress assessment—the field lacks comprehensive benchmarks encompassing diverse object categories, motion types, and conditioning modalities comparable to ImageNet~\cite{deng2009imagenet} on 2D tasks. \textit{Perceptual alignment gaps} between quantitative scores and human perception necessitate heavy reliance on manual evaluation, motivating efforts toward automated perception-aligned metrics~\cite{wang2024aligning}. Addressing these requires functionally-aware metrics, extended temporal assessment, comprehensive benchmarks, and improved perceptual alignment.

\begin{table*}[htbp!]
\caption{\textbf{A summary of representative methods}. Each method is characterized by its geometry, motion type, generative prior source, input condition, and training strategy. Motion: Articulation (ART), Deformation Field (DF), Tracking (TRK), Space-Time (ST), Per-frame (PF). \mrev{Prior source: foundation-model (FM; pretrained image/video/3D generative models), training-data (TD), input-only (Input), or language/vision-language model (LLM).}}
\vspace{-4pt}
\centering
\resizebox{0.98\textwidth}{!}{
\setlength{\tabcolsep}{12pt}
\begin{tabular}{lccccc}
\toprule
\textbf{Methods}  & \textbf{Geometry} & \textbf{Motion} & \mrev{\textbf{Prior}} & \textbf{Input Condition} & \textbf{Training Strategy} \\
\midrule

\rowcolor[HTML]{E3F2FD}
TextMesh4D~\cite{TextMesh4D} & Mesh & DF & FM & Text & Per-scene \\
\rowcolor[HTML]{F5FAFF}
V2M4~\cite{chen2025v2m4} & Mesh & DF & FM & Video & Per-scene \\
\rowcolor[HTML]{E3F2FD}
Puppeteer~\cite{song2025puppeteer} & Mesh & ART & FM+TD & Text/Mesh & Hybrid \\
\rowcolor[HTML]{F5FAFF}
AnimateAnyMesh~\cite{wu2025animateanymesh} & Mesh & DF & TD & Text+Mesh & Feed-forward \\
\rowcolor[HTML]{E3F2FD}
MagicArticulate~\cite{Song_2025_CVPR} & Mesh & ART & TD & Mesh & Feed-forward \\
\rowcolor[HTML]{F5FAFF}
DreamMesh4D~\cite{li2024dreammesh4d} & Mesh+Gaussian primitive & DF & FM & Video & Per-scene \\
\rowcolor[HTML]{E3F2FD}
RigAnything~\cite{liu2025riganything} & Mesh & ART & TD & Mesh & Feed-forward \\
\midrule

\rowcolor[HTML]{E8F5E9}
NeuralPCI~\cite{zheng2023neuralpci} & Point Cloud & DF & Input & Point Cloud & Per-scene \\
\rowcolor[HTML]{F5FBF6}
Cut3R~\cite{cut3r} & Point Cloud & PF & TD & Image & Feed-forward \\
\rowcolor[HTML]{E8F5E9}
Monst3R~\cite{zhang2024monst3r} & Point Cloud & PF+TRK & TD & Video & Feed-forward \\
\rowcolor[HTML]{F5FBF6}
St4RTrack~\cite{feng2025st4rtrack} & Point cloud & TRK & TD & Video & Feed-forward \\
\rowcolor[HTML]{E8F5E9}
PAPR-In-Motion~\cite{peng2024papr} & Point cloud & TRK & Input & Image & Per-scene \\
\rowcolor[HTML]{F5FBF6}
RPMNet~\cite{yan2020rpm} & Point cloud & DF & TD & Point Cloud & Feed-forward\\

\midrule

\rowcolor[HTML]{FFF3E0}
MAV3D~\cite{singer2023text} & NeRF & DF & FM & Text & Per-scene \\
\rowcolor[HTML]{FFFBF5}
4D-fy~\cite{bahmani20244dfy} & NeRF & DF & FM & Text & Per-scene \\
\rowcolor[HTML]{FFF3E0}
Consistent4D~\cite{jiang2023consistent4d} & NeRF & DF & FM+TD & Video & Per-scene \\
\rowcolor[HTML]{FFFBF5}
SV4D~\cite{xie_sv4d_2024} & NeRF & DF & FM+TD & Video & Hybrid \\
\rowcolor[HTML]{FFF3E0}
Dream-in-4D~\cite{zheng_unified_2024} & NeRF & DF & FM & Text/Image & Per-scene \\
\rowcolor[HTML]{FFFBF5}
Animate124~\cite{zhao2023animate124} & NeRF & DF & FM & Text+Image & Per-scene \\
\rowcolor[HTML]{FFF3E0}
4Diffusion~\cite{zhang_4diffusion_2024} & NeRF & DF & FM+TD & Video & Hybrid \\
\rowcolor[HTML]{FFFBF5}
V4D~\cite{gan2023v4d} & NeRF & ST & Input & Video & Per-Scene \\
\rowcolor[HTML]{FFF3E0}
TempInt~\cite{park2023temporal} & NeRF & ST & Input & Video & Per-Scene \\

\midrule

\rowcolor[HTML]{F3E5F5}
4DGen~\cite{yin_4dgen_2023} & Gaussian primitive & DF & FM & Video & Per-scene \\
\rowcolor[HTML]{FAF5FB}
4D-Rotor~\cite{duan20244d} & Gaussian primitive & ST & Input & Video & Per-scene \\
\rowcolor[HTML]{F3E5F5}
4D-GS~\cite{Wu_2024_CVPR} & Gaussian Primitive & ST/DF & Input & Video & Per-scene \\
\rowcolor[HTML]{FAF5FB}
Dynamic 3DGS~\cite{luiten2024dynamic} & Gaussian Primitive & ST/TRK & Input & Video & Per-scene \\
\rowcolor[HTML]{F3E5F5}
CAT4D~\cite{wu2025cat4d} & Gaussian primitive & DF & FM+TD & Video & Hybrid \\
\rowcolor[HTML]{FAF5FB}
DG4D~\cite{ren2023dreamgaussian4d} & Gaussian primitive & DF & FM & Image & Per-scene \\
\rowcolor[HTML]{F3E5F5}
4D-LRM~\cite{ma20254d} & Gaussian primitive & ST & TD & Few-Image & Feed-forward \\
\rowcolor[HTML]{FAF5FB}
L4GM~\cite{ren2024l4gm} & Gaussian primitive & PF & FM+TD & Video & Feed-forward \\
\rowcolor[HTML]{F3E5F5}
STAG4D~\cite{zeng2024stag4d} & Gaussian primitive & DF & FM & Text/Video & Per-scene \\
\rowcolor[HTML]{FAF5FB}
GenXD~\cite{zhao2024genxd} & Gaussian primitive & DF & FM+TD & Image & Hybrid \\
\rowcolor[HTML]{F3E5F5}
Mosca \cite{lei2024mosca} & Gaussian primitive & DF+TRK & Input & Video & Per-scene \\
\rowcolor[HTML]{FAF5FB}
Gaussian Marbles~\cite{stearns2024dynamic} & Gaussian primitive & TRK & Input & Video & Per-scene \\
\rowcolor[HTML]{F3E5F5}
In-2-4D~\cite{nag20252} & Gaussian primitive & DF & FM+LLM & Few Image & Per-scene \\
\rowcolor[HTML]{FAF5FB}
Free4D~\cite{liu2025free4d} & Gaussian primitive & DF & FM & Text+Image & Per-scene \\
\rowcolor[HTML]{F3E5F5}
EG4D~\cite{sun_eg4d_2024} & Gaussian primitive & DF & FM & Image & Per-scene \\
\rowcolor[HTML]{FAF5FB}
MVTokenFLow~\cite{huang2025mvtokenflow} & Gaussian primitive & DF+TRK & FM & Video & Per-scene \\
\rowcolor[HTML]{F3E5F5}
4Real~\cite{yu_4real_2024} & Gaussian primitive & DF & FM & Text & Per-scene \\
\rowcolor[HTML]{FAF5FB}
SC4D~\cite{wu_sc4d_2024} & Gaussian primitive & ART/DF & FM & Video & Per-scene \\

\midrule

\rowcolor[HTML]{E0F2F1}
PhysAvatar~\cite{PhysAavatar24} & Template/Mesh & ART/DF & Input & Video+Mesh & Per-scene \\
\rowcolor[HTML]{F2F9F9}
TADA~\cite{liao2024tada} & Template & ART/DF & FM+LLM & Text & Per-scene \\
\rowcolor[HTML]{E0F2F1}
ANYTOP~\cite{gat2025anytop} & Template & ART & TD & Skeleton & Feed-forward \\
\rowcolor[HTML]{F2F9F9}
Human3R~\cite{chen2025human3r} & Template/Point Cloud & ART & TD & Video & Feed-forward \\
\rowcolor[HTML]{E0F2F1}
MVP4D~\cite{taubner2025mvp4d} & Template/Gaussian Primitive & ART/DF & FM+TD & Image & Hybrid \\
\rowcolor[HTML]{F2F9F9}
Avatar Artist~\cite{liu2025avatarartist} & Template/NeRF & ART/DF & FM+TD & Image & Feed-forward \\
\rowcolor[HTML]{E0F2F1}
Disco4D~\cite{pang_disco4d_2024} & Template/Gaussian Primitive & ART/DF & FM & Image & Per-scene \\
\rowcolor[HTML]{F2F9F9}
CAP4D~\cite{taubner2025cap4d} & Template/Gaussian Primitives & ART/DF & FM+TD & Image & Hybrid \\
\rowcolor[HTML]{E0F2F1}
Vid2Avatar~\cite{guo2023vid2avatar} & Template/NeRF & ART/DF & Input & Video & Per-scene \\

\midrule

\rowcolor[HTML]{FFE0B2}
Paris~\cite{liu2023paris} & Part/NeRF & ART & Input & Image & Per-scene \\
\rowcolor[HTML]{FFF4E6}
ArticulatedGS~\cite{guo2025articulatedgs} & Part/Gaussian primitive & ART & Input & Image & Per-scene \\
\rowcolor[HTML]{FFE0B2}
ArtGS~\cite{liu2025artgs} & Part/Gaussian primitive & ART & Input & Image & Per-scene \\
\rowcolor[HTML]{FFF4E6}
SP4D~\cite{zhang2025sp4d} & Part/Mesh & ART & FM+TD & Video & Hybrid \\
\rowcolor[HTML]{FFE0B2}
ArticulateAnyMesh~\cite{qiu2025articulate} & Part/Mesh & ART & FM+LLM & Mesh & Per-scene \\
\rowcolor[HTML]{FFF4E6}
SINGAPO~\cite{liu2024singapo} & Part/Graph/Mesh & ART & TD+LLM & Image & Feed-forward \\
\rowcolor[HTML]{FFE0B2}
GeoPard~\cite{goyal2025geopard} & Part/Point Cloud & ART & TD & Point Cloud & Feed-forward \\
\rowcolor[HTML]{FFF4E6}
MeshArt~\cite{gao2025meshart} & Part/Mesh & ART & TD & -- & Feed-forward \\
\midrule

\rowcolor[HTML]{EDE7F6}
3DSG~\cite{3DDSG} & Graph & Scene Graph & Input & Stereo+IMU & Per-scene \\
\rowcolor[HTML]{F8F5FA}
4D-PSG~\cite{4DPanoSceneGraph} & Graph & Scene Graph & TD & Point Cloud & Feed-forward \\
\rowcolor[HTML]{EDE7F6}
PSG-4D-LLM\cite{wu2025learning} & Graph & Scene Graph & TD+LLM & Point Cloud & Feed-forward \\

\bottomrule
\end{tabular}}
\label{tab:4d_representations_reorganized}
\end{table*}

\section{Training Strategy}
\label{sec: training paradigm}

\begin{figure*}
    \centering
    \includegraphics[width=\linewidth]{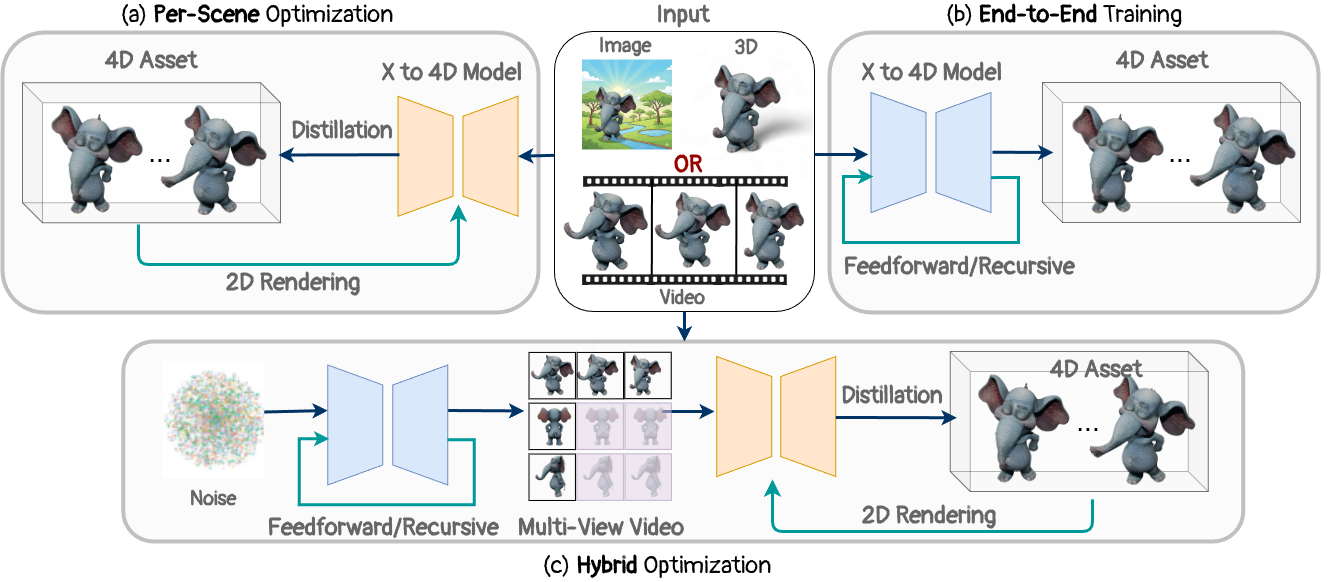}
\caption{Overview of three main training paradigms for 4D generation: (a) Per-scene optimization treats each scene as an individual optimization problem, leveraging information distilled from large foundation models (e.g., image and video diffusion models via score distillation sampling) to optimize a 4D representation; (b) End-to-end training employs feed-forward models trained on large-scale native 4D assets to learn strong data priors; (c) Hybrid optimization combines both approaches by training a customized feed-forward multi-view video generator, then using the generated videos to optimize individual scenes.
\vspace{-2em}}
\label{fig:training}
\end{figure*}

\mingz{The training strategies employed in 4D tasks reflect fundamental trade-offs between generalization, fidelity, and efficiency. For single-entity geometry and motion modeling, these strategies fall into three categories: per-scene optimization (\Cref{sec:perscene}), feed-forward models (\Cref{sec:feedforward}), and hybrid approaches that combine both (\Cref{sec:hybrid}). Training interaction models, however, introduces distinct challenges, namely coordinating multi-entity dynamics under physical constraints, that warrant separate treatment (\Cref{sec:interaction_training}). Building upon the geometric representations (\Cref{sec: geometry}) and motion modeling approaches (\Cref{sec: motion}) discussed above, we provide a comprehensive taxonomy of selected papers organized by these dimensions in~\Cref{tab:4d_representations_reorganized}.}
\mrev{Complementarily, \Cref{tab:interaction_methods} summarizes representative \emph{interaction}-centered methods along the same dimensions, tagging each by the interaction aspects it models---pose, contact, affordance, and physical regulation.}


\subsection{Per-Scene Optimization}
\label{sec:perscene}
Per-scene optimization treats each dynamic scene as an independent problem, directly optimizing model parameters (geometry, appearance, motion fields) from input observations without large-scale pre-training. This approach delivers high fidelity by dedicating computational resources to scene-specific refinement, operates without 4D training data by leveraging pre-trained 2D/video diffusion models~\cite{singer2023text,zeng2024stag4d}, and supports flexible scene-specific constraints with interpretable representations~\cite{zheng_unified_2024,wangshapemotion4dreconstructionsingle2024c}.

Below, we discuss some of the popular strategies used in per-scene optimization for both 4D generation and reconstruction tasks. 

\subsubsection{Data-driven Priors}

\textbf{Foundation model priors} provide implicit knowledge from large-scale datasets as substitutes for scarce 4D supervision. Multi-modal diffusion priors address complementary quality aspects: image diffusion~\cite{rombach2022high} for appearance detail, 3D-aware image diffusion~\cite{shi2023zero123++,long2023wonder3d} for spatial consistency, video diffusion~\cite{blattmann2023stable,singer2023makeavideo,guo2023animatediff} for temporal dynamics, and multi-view diffusion~\cite{shi2023mvdream,liu2023syncdreamer} for $360^{\circ}$ consistency. 4D-fy~\cite{bahmani20244dfy} alternates supervision across all three types, achieving 67\% user preference over MAV3D~\cite{singer2023text}. Recent work utilizes MLLMs/VLMs for 4D generation~\cite{zhou2025uni4d} and LLM reasoning for object composition~\cite{xu2024comp4d}.

\noindent\textbf{Auxiliary priors} such as depth offer geometric cues, optical flow offers motion cues, and semantic priors enforce region- or part-level consistency across time, thus stabilizing optimization, improving spatial-temporal coherence, and enhancing both geometric fidelity and perceptual realism in 4D tasks. Depth supervision improves joint learning of radiance and motion fields~\cite{caiNeuralSurfaceReconstructionDynamicScenes2022,wang2022learning}, particularly in outdoor scenes~\cite{wu2023mars,tonderski2024neurad,yang2023unisim}. Monocular depth estimation~\cite{ranftl2020towards,yang2024depth} provides geometric cues despite scale ambiguity~\cite{wangshapemotion4dreconstructionsingle2024c,liu2025modgs}. Optical flow models~\cite{xu2023unifying} supply dense inter-frame correspondences~\cite{yang2022banmo,liu2022devrf}, while visual foundation models like DINO~\cite{oquab2024dinov2} deliver semantic features consistent across views and time~\cite{yang2023emernerf,wu2023magicpony}. SA4D \cite{ji2024segment} lifts SAM~\cite{kirillov2023segment}-style 2D masks into 4D Gaussian representations by learning a temporal identity feature field. In dynamic scenes, object-level semantic segmentation provides valuable silhouettes or foreground masks that help localize dynamic objects and decompose scenes into static backgrounds and dynamic foregrounds~\cite{yanNerfdsNeuralRadianceFieldsDynamic2023,li2024sadg}
. For object articulation~\cite{guo2025articulatedgs}, instead of using foundational model priors, it uses physics-based optimization \cite{igarashi2005rigid} to learn motion parameters. 

\subsubsection{Design Choices}
Converting 2D or textual inputs into 3D (and subsequently 4D) structure depends heavily on how the input modality is encoded and lifted. Here we discuss some of the design choices that have influenced 4D literature in the realm of per-scene optimization.
 
\noindent\textbf{3D lifting models} differ in how they convert 2D/textual inputs into 3D structure. Multi-view generation approaches like Zero-1-to-3~\cite{liu2023zero} and MVDream~\cite{shi2023mvdream} for objects, as well as ViewCrafter~\cite{yu2024viewcrafter} and SEVA~\cite{zhou2025stable} for scenes, synthesize novel views and reconstruct structure through NeRF or Gaussian Splatting. This has recently been improved by multi-view video generators~\cite{voleti2024sv3d} for better geometry quality. Direct feed-forward methods~\cite{xiang2025structured,tang2024lgm,jun2023shap} infer 3D structure from single images via learned mappings. Multi-view synthesis ensures stronger geometric consistency; feed-forward prediction offers faster, generalizable inference at reduced fidelity cost.

\noindent\textbf{Disentanglement methods} decouple static and dynamic components for easier optimization. Two-stage approaches dominate: Consistent4D~\cite{jiang2023consistent4d}, STAG4D~\cite{zeng2024stag4d}, and 4D-fy~\cite{bahmani20244dfy} optimize static 3D assets before adding motion through deformation learning. MAV3D~\cite{singer2023text} uses three stages: static, dynamic, and super-resolution.

\noindent\textbf{Deformation fields} model motion patterns using MLPs~\cite{bae2024pergaussian,huang2024sc,luiten2024dynamic}, spatial-temporal planes~\cite{duisterhof2023deformgs,lu2024dn4dgs,Wu_2024_CVPR}, polynomial functions~\cite{li2024spacetime}, Fourier series~\cite{katsumata2024compact}, or combinations~\cite{lin2024gaussian}. These methods directly learn motion patterns from independent time inputs, neglecting cross-time relationships. Recent work~\cite{jiang2024timeformer} adaptively learns temporal relationships via transformer attention. Most methods optimize deformation frame-by-frame from multi-view videos ~\cite{yao2025sv4d,zeng2024stag4d}, while articulation-based methods~\cite{liu2025artgs,guo2025articulatedgs} estimate motion from heuristics or input geometry.

\subsubsection{Optimization Objectives}

\textbf{Score Distillation Sampling (SDS)} distills prior knowledge for 4D generation. Originating with DreamFusion~\cite{poole2022dreamfusion}, SDS iteratively optimizes representations by rendering views, adding noise, denoising with diffusion models, and computing gradients:
\begin{equation}
   \nabla_{\theta} L_{\mathrm{SDS}} = \mathbb{E}_{t,\epsilon}\!\left[w(t)\big(\epsilon_{\phi}(x_t,t,c) - \epsilon\big)\nabla_{\theta} x_t\right]
\end{equation}
This distills 2D priors into 3D/4D space without 3D training data~\cite{alldieck2024score,poole2022dreamfusion}, typically driving deformation fields from generated multi-view videos~\cite{bahmani2025lyra,zeng2024stag4d}.

\noindent\textbf{Reconstruction loss} provides direct supervision when reference views exist. Shape of Motion \cite{wangshapemotion4dreconstructionsingle2024c} optimizes photometric consistency between rendered and captured views, depth supervision from monocular depth priors, and 2D tracking consistency with long-range point tracks. The photometric reconstruction loss is implemented as L2 distance between the generated/reconstructed source view with the target view as follows: 
\begin{equation}
    \mathcal{L}_\textrm{rec} = \frac{1}{\mathcal{T}}\sum_{\tau=1}^{\mathcal{T}}||f(\tau,o_\textrm{Ref}) - I^\tau_\textrm{Ref}||^2_2,
\end{equation}
where $\mathcal{T}$ is the frame count, $f(.)$ is the rendering function and $o_\textrm{Ref}$ is the reference camera. For object centric scenes \cite{jiang2023consistent4d}, a separate foreground mask loss \cite{wangshapemotion4dreconstructionsingle2024c,zeng2024stag4d} is also estimated. 
These reconstruction losses complement or replace SDS when ground truth video supervision is available. These losses primarily enhance per-frame appearance and reduce floating artifacts.

\noindent \textbf{Regularization losses} ensure physical plausibility. Temporal and spatial smoothness use Total Variation loss~\cite{lombardiNeuralVolumesLearningDynamicRenderable2019,fridovich2023k,huang2024textit} and Laplacian regularization~\cite{habermann2021real}. Structural integrity is preserved via As-Rigid-As-Possible constraints~\cite{igarashi2005rigid}, isometric loss, and divergence loss. ArtGS~\cite{liu2025artgs} uses Chamfer distance for geometry preservation during articulation.

\subsubsection{Open Challenges}
Optimization-based 4D methods achieve high visual fidelity, but remain impractical for large-scale use due to computational expense, instability, and slow convergence, limiting scalability and motion complexity in real-world applications. Key limitations include:

\noindent \textbf{Compute cost.} These methods are extremely time-consuming, with \cite{zeng2024stag4d} and \cite{jiang2023consistent4d} requiring around 2 hours per scene and \cite{singer2023text} or \cite{bahmani20244dfy} taking several to $10+$ hours. This makes them $10-1000$x slower than feed-forward models, preventing real-time or interactive use. They also demand high-end GPUs like V100/A100 and large memory budgets (at least 24 GB GPU RAM), posing significant hardware barriers.

\noindent \textbf{Scalability.} Each scene must be optimized independently, eliminating the possibility of batch or parallel generation. Such slow, scene-specific pipelines are impractical for large-scale or commercial deployment, where generating hundreds of 4D assets for gaming, film, or AR/VR would be prohibitively costly. Moreover, learned representations do not generalize across scenes. 

\noindent \textbf{Limited motion complexity.} Current pipelines perform well for simple, object-scale motions but fail to capture large, scene-level dynamics, non-rigid deformations, or topological changes. Realistic physics is rarely enforced, and long-range motion across extended sequences remains unsolved. Modeling the appearance or disappearance of objects over time remains particularly challenging for continuous representations.

\begin{table*}[htbp!]
\caption{\mrev{\textbf{A summary of representative interaction-centered methods}. The \textbf{Interaction} column tags the interaction modelling strategies \iPose~pose modelling, \iCont~contact modelling, \iAfd~affordance modelling, \iPReg~physical regulation. The \textbf{Task} column denotes the interaction scope: HOI (human-object), HSI (human-scene), HHI (human-human), Hand-O (hand-object), and others.}}
\vspace{-4pt}
\centering
\resizebox{\textwidth}{!}{
\setlength{\tabcolsep}{9pt}
\begin{tabular}{lccccccc}
\toprule
\textbf{Methods}  & \textbf{Geometry} & \textbf{Motion} & \textbf{Prior} & \textbf{Input Condition} & \textbf{Training Strategy} & \textbf{Interaction} & \textbf{Task} \\
\midrule

\rowcolor[HTML]{E0F2F1}
ZeroHSI~\cite{li2024zerohsi} & Template/Gaussian primitive & ART & FM & Text+Scene & Per-scene & \iPose~\iPReg & HSI \\
\rowcolor[HTML]{F2F9F9}
MANUS~\cite{pokhariya2024manus} & Template/Gaussian primitive & ART & Input & Multi-view & Per-scene & \iPose~\iCont & Hand-O \\
\rowcolor[HTML]{E0F2F1}
Open3DHOI~\cite{wen2025reconstructing} & Template/Gaussian primitive & ART & FM+Human & Image & Per-scene & \iPose~\iCont~\iPReg & HOI \\
\rowcolor[HTML]{F2F9F9}
AvatarGO~\cite{cao_avatargo_2024} & Template/Gaussian primitive & ART/DF & FM+LLM & Text & Per-scene & \iPose~\iPReg & HOI \\
\rowcolor[HTML]{E0F2F1}
HOSNeRF~\cite{liu2023hosnerf} & Template/NeRF & ART/DF & Input & Video & Per-scene & \iPose & HOI/HSI \\
\rowcolor[HTML]{F2F9F9}
HandNeRF~\cite{guo2023handnerf} & Template/NeRF & ART/DF & Input & Multi-view & Per-scene & \iPose & Hand-Hand \\
\rowcolor[HTML]{E0F2F1}
BIGS~\cite{on2025bigs} & Template/Gaussian primitive & ART & FM+Input & Video & Per-scene & \iPose~\iPReg & Hand-O \\
\rowcolor[HTML]{F2F9F9}
Neural Descriptor Fields~\cite{NeuralDescriptorFields} & Point cloud & -- & TD & Point Cloud & Feed-forward & \iPose & OOI \\
\rowcolor[HTML]{E0F2F1}
InterGen~\cite{Intergen} & Template & ART & TD & Text & Feed-forward & \iPose & HHI \\
\rowcolor[HTML]{F2F9F9}
POSA~\cite{POSA} & Template & -- & TD & Mesh+Scene & Feed-forward & \iPose~\iCont~\iAfd~\iPReg & HSI \\
\rowcolor[HTML]{E0F2F1}
GanHand~\cite{corona2020ganhand} & Template & -- & TD & Image & Feed-forward & \iPose~\iAfd~\iPReg & Hand-O \\
\rowcolor[HTML]{F2F9F9}
3D AffordanceNet~\cite{3DAffordanceNet} & Point cloud & -- & TD & Point Cloud & Feed-forward & \iAfd & Affordance prediction \\
\rowcolor[HTML]{E0F2F1}
AffordanceLLM~\cite{AffordanceLLM} & Image & -- & TD+LLM & Image+Text & Feed-forward & \iAfd & Affordance prediction \\
\rowcolor[HTML]{F2F9F9}
NIFTY~\cite{kulkarni2024nifty} & Template & ART & TD & Object & Feed-forward & \iPose~\iCont~\iAfd & HOI \\
\rowcolor[HTML]{E0F2F1}
InterDiff~\cite{xu2023interdiff} & Template/Mesh & ART & TD & Motion & Feed-forward & \iPose~\iPReg & HOI \\
\rowcolor[HTML]{F2F9F9}
FORCE~\cite{FORCEIntuitivePhysics} & Template & ART & TD & Action+Object & Feed-forward & \iPose~\iCont~\iPReg & HOI \\

\bottomrule
\end{tabular}}
\label{tab:interaction_methods}
\end{table*}

\subsection{End-to-End Training}
\label{sec:feedforward}
Feed-forward models learn global mappings from 2D observations to 4D representations using large pre-trained networks. Once trained, these models infer novel scenes in single forward passes, enabling real-time generation (e.g., L4GM~\cite{ren2024l4gm} at 0.3s, 4DGT~\cite{xu20254dgt} at 25 ms/frame) and cross-dataset scalability with learned spatiotemporal priors that handle ambiguity and sparse views, and provide deterministic inference with stable quality and fixed runtime regardless of scene complexity.

\subsubsection{Data-driven Priors}
\textbf{Motion priors.} Some methods leverage video diffusion models for joint motion-appearance learning. 4DNeX~\cite{chen20254dnex} integrates RGB and 3D sequences using DiT architectures. Geo4D~\cite{jiang2025geo4d} extends image-to-video diffusion to output explicit geometry with point maps. Other works, such as AnimateAnyMesh~\cite{wu2025animateanymesh} and Motion 3-to-4~\cite{chen2026motion} directly learn motion prior from large animated 3D assets.

\noindent \textbf{Geometry priors.}
Large 3D backbones are widely adopted to anchor geometry priors: L4GM builds on LGM~\cite{tang2024lgm}, 4D-LRM~\cite{ma20254d} extends LRM~\cite{hong2023lrm}, and Forge4D~\cite{hu2025forge4d} adopts VGGT. MonST3R~\cite{zhang2024monst3r} and Geo4D~\cite{jiang2025geo4d} employ DUSt3R encoders.
Human-object interaction methods~\cite{wen2025reconstructing,lee2023im2hands,lee2024interhandgen} anchor on predefined template geometries~\cite{pavlakos2019expressive,MANO:SIGGRAPHASIA:2017}. Together, these priors transfer pre-trained 3D spatial understanding to dynamic 4D tasks, achieving high-quality results with minimal 4D supervision.

\subsubsection{Design Choices}
The fundamental design of a 4D model is defined by how it represents 3D space and captures the evolution of that space over time.

\noindent \textbf{3D lifting models.} 
Most recent works adopt 3D Gaussian Splatting (3DGS)~\cite{kerbl3dgaussians} for its balance of realism, efficiency, and differentiability. Two main extensions adapt it to 4D:

\noindent\textit{Unified 4D Representations:}  
Methods like 4DGT~\cite{xu20254dgt} embed temporal attributes (e.g., velocity, lifespan) directly into each Gaussian, while 4D-LRM~\cite{ma20254d} models space–time jointly via anisotropic 4D Gaussians for fast reconstruction.

\noindent\textit{\mingz{Sequential or Deformable Fields:}}  
Other approaches construct sequential 3DGS scenes linked by interpolation~\cite{ren2024l4gm}, continuous flow~\cite{Lu2024DrivingRecon} \mingz{or conditioned on timestamps~\cite{liang2024feed}. Beyond GS, ~\cite{wu2025animateanymesh,chen2026motion} handle mesh animations through vertex displacement maps.}
Overall, current trends favor explicit, geometry-aware, and temporally coherent 3D lifting frameworks that unify appearance and motion in a feed-forward manner.

\noindent\textbf{Motion models} have evolved beyond simple interpolation. Explicit motion prediction methods~\cite{lin2025moviesmotionaware4ddynamic,hu2025forge4d} use transformer heads to predict dense 3D motion fields or per-pixel displacements. Temporal transformers~\cite{ren2024l4gm} incorporate attention for cross-frame dependencies, enabling implicit learning of complex interactions—a popular choice in human-object interaction~\cite{lee2024interhandgen,xu2023actformer}, recently replaced by diffusion methods~\cite{ron2025hoidini,song2024hoianimator}. 
Together, these paradigms transition from per-frame fitting, \mingz{in-between bridging }to learned feed-forward representations, improving temporal continuity and scalability.

\subsubsection{Optimization Objectives}

\noindent\textbf{Reconstruction losses} optimize photometric consistency using MSE/L1, SSIM, and LPIPS for perceptual sharpness~\cite{ren2024l4gm,ma20254d}. Extensions include segmentation-aware L1~\cite{Lu2024DrivingRecon} \mingz{for static-dynamic decoupling}. These losses ensure visual accuracy and serve as universal reconstruction anchors across Gaussian, mesh, and diffusion systems.

\noindent\textbf{Diffusion-based losses} minimize noise prediction errors in latent video space~\cite{chen20254dnex,jiang2025geo4d}, predicting $\epsilon$ conditioned on frames or camera embeddings. 
AnimateAnyMesh~\cite{wu2025animateanymesh} adopts flow-matching objectives for continuous-time supervision.  These losses align generated sequences with learned diffusion manifolds, ensuring smooth motion and appearance trajectories.

\noindent \textbf{Geometry losses} ensure metric realism. 4DGT~\cite{xu20254dgt} uses depth~\cite{yang2024depth} and normal~\cite{ye2024stablenormal} supervision. DrivingRecon~\cite{Lu2024DrivingRecon} employs a DepthNet for depth integration. MonST3R~\cite{zhang2024monst3r} imposes pointmap alignment and flow-projection consistency, while Geo4D~\cite{jiang2025geo4d} merges multiple geometric terms. 
Interaction methods~\cite{lee2024interhandgen,ron2025hoidini,xu2023interdiff} employ physics-based losses to penalize interpenetration, enforce contact constraints, and regularize trajectories.

\subsubsection{Open Challenges}
End-to-end training enables joint learning of geometry, appearance, and motion within a unified framework but introduces limitations in data requirements, generalizability, and architectural flexibility. We discuss some of the drawbacks below:

\noindent\textit{Training data.} Feed-forward models require massive-scale 4D data that is costly to acquire. Existing methods~\cite{xu20254dgt,ren2024l4gm,ma20254d,chen20254dnex} train on mixed 3D datasets~\cite{zhou2018stereo,deitke2024objaverse} or limited explicit 4D datasets~\cite{grauman2024ego,jin2024stereo4d}. L4GM~\cite{ren2024l4gm} requires 12M synthetic videos (300M frames), while 4DGT~\cite{xu20254dgt} uses ~5K real-world videos. Training demands days on 64–128 GPUs with complex multi-stage optimization, making academic-scale training infeasible due to computational and storage costs.

\noindent \textit{Limited generalizability.} Models trained on narrow distributions struggle beyond training conditions. L4GM~\cite{ren2024l4gm} assumes fixed camera elevation and fails on egocentric or occluded motion; 4DGT~\cite{xu20254dgt} performs poorly on new camera types; 4DNeX~\cite{chen20254dnex} degrades under occlusion or lighting changes. Domain shift from synthetic to real scenes remains critical—models trained on Objaverse-Dy~\cite{yao2025sv4d} fail to generalize to real-world scenarios, limiting practical applicability.

\noindent \textit{Lack of flexibility.} Fixed architectural designs limit adaptability. L4GM~\cite{ren2024l4gm} directly predicts Gaussian splats, tightly coupling the network to this representation—switching to meshes or implicit fields requires full retraining. Similarly, 4DGT~\cite{xu20254dgt} and St4RTrack~\cite{feng2025st4rtrack} hard-code 4D Gaussians or pointmap assumptions. This contrasts with per-scene optimization frameworks that can swap representations or regularizations without retraining, offering greater experimental flexibility.

\subsection{Hybrid Optimization}
\label{sec:hybrid}
Hybrid methods combine the strengths of the two extremes. Typically, a feed-forward backbone provides a strong initialization or prior, which is then refined through scene-specific optimization. This two-stage paradigm balances reconstruction quality and inference speed—leveraging global priors for generalization and local optimization for fidelity.

\noindent \textbf{Two-stage paradigm.} 
A distinct family of 4D methods adopts a \textit{two-stage design}: Stage~1 performs feed-forward multi-view or video generation, while Stage~2 refines or reconstructs explicit 4D geometry. 
This combines the scalability of diffusion-based appearance synthesis with the precision of geometric optimization. 
CAT4D~\cite{wu2025cat4d} first generates view-consistent videos using a multi-view diffusion model (CAT3D + Lumiere), then reconstructs 4D motion via deformable Gaussian splatting. 
SV4D~\cite{xie_sv4d_2024} and SV4D~2.0~\cite{yao2025sv4d} similarly use Stable Video Diffusion for multi-view synthesis followed by dynamic NeRF optimization. 
Splat4D~\cite{yin2025splat4d} predicts coarse 3D Gaussians before refining temporal coherence with DynamiCrafter, while DreamArt~\cite{lu2025dreamart} synthesizes articulated motion via video diffusion and reconstructs textured meshes through dual-quaternion optimization. Overall, this two-stage \textit{feed-forward + optimization} framework balances generative realism and geometric consistency, yielding high-fidelity and temporally stable 4D reconstructions.

\noindent \textbf{SDS-Free optimization.} \mingz{A notable trend in hybrid 4D methods is the shift away from score distillation sampling (SDS) to direct photometric, geometric, and temporal supervision—enabling Stage~2 optimization to remain lightweight and fast. This trend is also broadly observed across per-scene and feed-forward paradigms.
Hybrid methods like AR4D~\cite{zhu2025ar4d}, Splat4D~\cite{yin2025splat4d}, SV4D~\cite{xie_sv4d_2024}, and SV4D 2.0~\cite{yao2025sv4d} rely on flow alignment, uncertainty minimization, and SDS-free NeRF optimization, respectively.} Per-scene methods like EG4D~\cite{sun_eg4d_2024} and In-2-4D~\cite{nag20252} optimize deformation fields using photometric and geometric losses. Feed-forward methods such as BTimer~\cite{liang2024feed}, 4DGT~\cite{xu20254dgt}, DrivingRecon~\cite{Lu2024DrivingRecon}, MoVies~\cite{lin2025moviesmotionaware4ddynamic}, and Forge4D~\cite{hu2025forge4d} leverage depth, flow, photometric, and perceptual losses, making SDS unnecessary.

\subsection{\mingz{Interaction Training}}
\label{sec:interaction_training}

\mingz{Interaction training aims to generate physically coherent motion sequences for multiple entities from conditioning signals such as text~\cite{diller2024cg}, audio~\cite{siyao2024duolando}, objects~\cite{li2023object}, scenes~\cite{hassan2021stochastic}, or action classes~\cite{xu2023actformer}. Human entities are predominantly modelled with template-based representations—either fully-parametric meshes or skeletons—sometimes augmented or replaced by neural representations such as NeRF and Gaussian Splatting for photorealism, whereas object entities span both structured and unstructured representations. The central challenge lies in capturing not only individual motions but also inter-entity relationships. \gmark{Existing approaches broadly divide into data-driven and reward-driven paradigms; we briefly overview both below and refer readers to~\cite{sui2026survey} for in-depth discussion.}}

\subsubsection{Data-Driven}
\mingz{Data-driven methods learn to map input conditions to interaction motion sequences from captured data priors (see ~\Cref{subsec: interaction}). Architectures have evolved from RNNs~\cite{kundu2020cross} and GANs~\cite{xu2023actformer} through VAEs~\cite{GOALGrasping, wu2022saga} to the now-dominant diffusion models~\cite{Intergen, li2023object} and transformers~\cite{ghosh2025duetgen}. Training can be further regularized by contact~\cite{wang2025end}, affordance~\cite{zhang2025openhoi}, collision~\cite{zhang2020generating}, and physical priors~\cite{FORCEIntuitivePhysics}. While this paradigm produces natural motions grounded in real human data, physical plausibility remains only implicitly enforced, GOAL~\cite{GOALGrasping} and SAGA~\cite{wu2022saga} resort to post-hoc optimization to reduce penetration, and HOI-Dyn~\cite{lin2025hoidyn} introduces explicit driver--responder synchronization via a transformer to improve inter-entity coherence.}

\subsubsection{Reward-Driven}
\mingz{Reward-driven methods instead train agents via reinforcement learning coupled with a physics simulator, making physical plausibility an intrinsic guarantee rather than a learned prior. The central design problem is the reward function. \cite{hassan2023synthesizing} combines a task-completion reward with a learned realism reward derived from motion data; \cite{wang2025physhsi} extends this design to real humanoid robots. SMP~\cite{mu2025smp} repurposes a pretrained motion diffusion model as a frozen, task-agnostic reward, enabling a single prior to guide diverse policies. A key limitation is that most methods train separate per-task policies. \cite{xu2025intermimic} addresses this via teacher--student distillation across diverse HOI datasets, correcting MoCap artifacts through physics simulation. Nevertheless, reward-driven approaches remain constrained by per-task reward engineering and reliance on reference motions, limiting generalization.}

\subsubsection{Open Challenges}

\mingz{\noindent\textit{Realism.} Template-based body representations cannot capture soft deformation upon contact, limiting realistic grasping and collision-free close interaction. MoCap artifacts such as floating contacts further degrade supervision quality across both paradigms.}

\noindent\mingz{\textit{Generalization.} Neither paradigm generalizes well beyond its training distribution. LLM-guided methods partially address this by decomposing open-vocabulary instructions into structured sub-goals for low-level controllers~\cite{xiao2023unified, wang2025sims, cao_avatargo_2024}, yet LLM-generated plans remain physically inconsistent. Tighter integration of LLM semantic reasoning with physics-based execution is a key open direction.}


\section{Representation Comparison and Trade-off}
\label{sec:rep_tradeoff}

We compare representations covered so far across seven key metrics; see below.
\Cref{tab:representation_properties} outlines the fundamental trade-offs.
\mrev{Alongside these properties, the table summarizes the 4D tasks each representation is most commonly chosen for in the literature. We consider seven recurring tasks: dynamic novel-view synthesis (\iNVS,~\cite{bahmani20244dfy, wu2025cat4d}), text/image/video/3D-to-4D generation (\iGEN,~\cite{chen2025v2m4, zhao2023animate124, nag20252, liu2024singapo}), articulation (\iART,~\cite{song2025puppeteer, liu2024cage, goyal2025geopard, SMPL:2015}), human and interaction modeling (\iHUM,~\cite{icsik2023humanrf,yuan2024gavatar,wu2025learning, hassan2021stochastic}), large unbounded scenes (\iLRG,~\cite{wang2025pi, li2023dynibar, lei2024mosca, ost2021neural}), feed-forward / efficient inference (\iFFW,~\cite{wu2025animateanymesh, cut3r, ma20254d}), and physics-grounded modelling (\iPHY,~\cite{qiu2023rec, xie2024physgaussian,PhysAavatar24}).}


\begin{table*}[t]
\centering
\caption{Comparison of 4D geometric representation properties across seven key dimensions relevant to dynamic scene modelling,
\mrev{and the 4D tasks each representation is most commonly chosen for}.
\mrev{\textbf{Task icons:}\;
\iNVS~novel-view synthesis,\;
\iGEN~4D generation,\;
\iART~articulation,\;
\iHUM~human / interaction,\;
\iLRG~large unbounded scenes,\;
\iFFW~feed-forward inference,\;
\iPHY~physics-grounded modelling.}}
\label{tab:representation_properties}
\resizebox{\textwidth}{!}{%
\begin{tabular}{@{}lccccccccc@{}}
\toprule
\textbf{Representation} &
\textbf{Motion Representation} &
\textbf{Visual Fidelity} &
\textbf{Scalability} &
\textbf{Temporal Consistency} &
\textbf{Topology Handling} &
\textbf{Editability} &
\textbf{Generalization} &
\textbf{Efficiency} &
\textbf{\mrev{Suited tasks}} \\
\midrule
\rowcolor{gray!10}
\textbf{Mesh} &
Vertex displacement &
Medium &
Medium &
High &
\textcolor{red}{\ding{55}} Fixed &
Medium &
High &
Low &
\iGEN~\iART~\iFFW~\iPHY \\
\rowcolor{gray!10}
& Skinning weights & & & (explicit tracking) & connectivity & & & & \\
\addlinespace
\rowcolor{blue!5}
\textbf{Point Cloud} &
Scene flow &
High &
Excellent &
Low &
\textcolor{green}{\ding{51}} Natural &
Medium &
High &
High &
\iLRG~\iFFW \\
\rowcolor{blue!5}
& vectors & & & (no structure) & (add/remove points) & & & & \\
\addlinespace
\rowcolor{gray!10}
\textbf{NeRF} &
Deformation field &
Very High &
High &
Medium &
\textcolor{green}{\ding{51}} Implicit &
High &
Medium &
Low &
\iNVS~\iGEN~\iHUM~\iLRG \\
\rowcolor{gray!10}
& Time-conditioned MLP & & & & (no topology) & &  & & \\
\addlinespace
\rowcolor{blue!5}
\textbf{Gaussian Splatting} &
Space-Time Gaussian &
Very High &
High &
Medium &
\textcolor{green}{\ding{51}} Adaptive &
High &
High &
High &
\iNVS~\iGEN~\iHUM~\iLRG~\iFFW~\iPHY \\
\rowcolor{blue!5}
& Deformation field & & & (Gaussian tracking) & (splat birth/death) & & & & \\
\addlinespace
\rowcolor{gray!10}
\textbf{Graph} &
Edge relationships &
Medium &
High &
Very High &
\textcolor{green}{\ding{51}} Dynamic &
Very High &
Medium &
Medium &
\iART~\iHUM~\iLRG \\
\rowcolor{gray!10}
& Node attributes & & & (node tracking) & (edge operations) & & (structure transfer) & & \\
\addlinespace
\rowcolor{blue!5}
\textbf{Part} &
Articulation &
Medium &
Medium &
Very High &
\textcolor{orange}{\ding{51}} Limited &
Very High &
Excellent &
Medium &
\iGEN~\iART~\iPHY \\
\rowcolor{blue!5}
& & & & (parametric) & (fixed primitive set) & & (within category) & & \\
\addlinespace
\rowcolor{gray!10}
\textbf{Template} &
Deformation basis &
Medium &
High &
Very High &
\textcolor{red}{\ding{55}} Fixed &
Very High &
Excellent &
Medium &
\iART~\iHUM~\iPHY \\
\rowcolor{gray!10}
& Articulation & & & (correspondence) & (topology-preserving) & & (within category) & & \\
\bottomrule
\end{tabular}
}
\end{table*}

\noindent\textbf{Visual Fidelity} (\textit{Quality of reconstructed appearance, including photorealism and geometric detail preservation}). Volumetric rendering (NeRF~\cite{park2021hypernerf}, 3DGS~\cite{lei2024mosca}) achieves the highest fidelity through view-dependent effects and continuous radiance modeling. Point clouds~\cite{cut3r} and meshes~\cite{zhao2025hunyuan3d} follow: point clouds directly use input pixel colors despite discrete surfaces, while meshes provide continuous geometry but require PBR modeling for photorealism. Structured representations (graph, part, template~\cite{liu2024singapo,liao2024tada}) exhibit medium fidelity as appearance depends on node or part attributes rather than direct modeling. 

\noindent\textbf{Scalability} (\textit{Quality and efficiency when extending from simple objects to large, complex scenes with multiple entities}). Point clouds excel at large-scale scenarios (autonomous driving~\cite{caesar2020nuscenes}, urban reconstruction~\cite{lin2022capturing}), while NeRF~\cite{attal2023hyperreel}, Gaussian Splatting~\cite{zhou2024drivinggaussian}, and graphs handle unbounded multi-entity scenes through volumetric fusion and explicit modeling. In contrast, mesh, part, and template methods face constraints from connectivity complexity and category-specific design. 

\noindent\textbf{Temporal Consistency} (\textit{Motion smoothness and coherence, i.e., absence of flickering, jittering, and discontinuity}). Structured methods (graph, part, template~\cite{zhang2025sp4d,peng2021neural}) achieve the highest consistency through explicit functional motion modeling, while meshes~\cite{wu2025animateanymesh} maintain high consistency via vertex tracking. NeRF~\cite{pumarola2021d} and 3DGS~\cite{bahmani20244dfy} exhibit moderate consistency as learned deformation fields require temporal regularization. Point clouds demonstrate variable performance: per-frame reconstruction with tracking~\cite{feng2025st4rtrack} achieves consistency, but native 4D sequences suffer from lack of explicit correspondence. \saura{Tracking-based motion representations can provide better and efficient temporal consistency in a zero-shot manner \cite{wangshapemotion4dreconstructionsingle2024c}, but may introduce errors when projecting  2D priors into 3D.}

\noindent\textbf{Topology Handling} (\textit{Topological changes during motion: splitting, merging, and appearance/disappearance of structures}). Implicit and adaptive methods (NeRF, Gaussian Splatting, point clouds, graphs) naturally accommodate topological changes through volumetric representation or dynamic structure modification. In contrast, fixed-connectivity approaches (mesh, template, part) require consistent topology throughout sequences. 

\noindent\textbf{Editability} (\textit{Ease of manipulation and control, including both semantic-level and fine-grained modifications}). Parametric representations (graph, part, template) achieve the highest editability through intuitive semantic controls, while explicit geometry (mesh, point clouds) provides straightforward but limited manipulation. Learned methods (NeRF, 3DGS) offer powerful editing capabilities but require less intuitive latent space manipulation.

\noindent\textbf{Generalization} (\textit{Transferability to unseen scenes/objects without  per-instance optimization or retraining}). Category-level methods (part, template) achieve excellent generalization, while the rest results in moderate cross-instance capability when trained on large-scale data~\cite{xiang2025structured,wang2025vggt,hong2023lrm, tang2024lgm}. \saura{Deformation-field motion representations~\cite{Li_2023_CVPR,yao2024dynosurf} offer flexible optimization over diverse inputs and enable stronger motion control. Being learning-based, they generalize better than zero-shot tracking, which is often prone to 2D-to-3D projection errors.}

\noindent\textbf{Efficiency} (\textit{Motion-coupled inference speed}). Point clouds~\cite{cut3r} and Gaussian Splatting with space-time motion~\cite{duan20244d} achieve the highest efficiency: point clouds through per-frame reconstruction without temporal dependencies, and space-time Gaussians by baking motion directly into the representation. Graph~\cite{fischermultilevelneuralscenegraphsdynamic2024}, template~\cite{pang_disco4d_2024} and part-based~\cite{goyal2025geopard} methods follow, benefiting from their compact parametric structure and inherent motion coupling (e.g., node motion or articulation). Methods requiring per-frame deformation evaluation—including mesh with vertex displacement~\cite{TextMesh4D}, NeRF~\cite{bahmani20244dfy} and 3D Gaussians~\cite{ling2024align} with learned deformations—are less efficient due to the computational overhead of temporal motion decoding.
\vspace{-1em}
\section{Conclusion, Emerging Trends, and Future Directions}
\label{sec: Conclusion}

We provide a representation-centric synthesis of recent advances in 4D modeling, highlighting how the choice of representation (e.g., structured vs.~unstructured) fundamentally shapes the design, capability, and limitation of methods to reconstruct and generate 4D content. By examining the interplay between representation, motion type, and temporal dynamics, as well as the associated datasets, metrics, and benchmarks, we reveal the underlying trade-offs that govern efficiency, fidelity, and generalization. This unified perspective not only bridges prior efforts focused on specific techniques, but also establishes a conceptual framework to guide future research toward more principled and scalable approaches to capturing, understanding, and synthesizing motions and interactions.

We conclude the survey by discussing emerging trends or paradigm shifts, and open challenges to stimulate future work.

\noindent\textbf{Trend \#1: Feed-Forward Reconstruction.} The transition from per-scene optimization to feed-forward inference represents a fundamental shift in computational efficiency. Traditional approaches requiring hours of optimization per scene are being replaced by Large Reconstruction Models (LRMs) for 4D through a single forward pass~\cite{ren2024l4gm,chen20254dnex}, achieving 100-1000× speedups. This trend mirrors similar transitions witnessed in 2D~\cite{rombach2022high} and 3D~\cite{zhang2024clay} generation, establishing feed-forward architectures as a dominant paradigm across dimensional barriers.

\noindent\textbf{Trend \#2: Hybrid Generation-Reconstruction Pipelines.} The boundary between the two has become increasingly blurred, with state-of-the-art methods employing generative models such as ``data amplifiers" to overcome sparse-view limitations. Multi-stage pipelines first employ diffusion models to synthesize missing multi-view observations from monocular inputs, then apply optimization-based reconstruction to the generated data~\cite{wu2025cat4d,zhang2025sp4d}. This paradigm addresses fundamental data scarcity through generative priors rather than improved capture hardware, but generation errors may be unavoidable and propagate into the reconstruction. 

\noindent\textbf{Trend \#3: Integrating World Knowledge.} Capturing 4D contents with real-world fidelity requires incorporating high-level semantic understanding and physical plausibility. The use of LLMs has become prevalent for multi-modal reasoning and common-sense guidance in 4D tasks~\cite{wu2025animateanymesh,liu2024singapo}, encoding abstract knowledge that would be impractical to specify as mathematical inductive biases or require prohibitive amounts of training data to learn. Complementarily, physics-based constraints are being integrated through differentiable physics losses~\cite{lu2025humoto} or reconstruct-then-simulate pipelines~\cite{xie2024physgaussian,zhang2024physdreamer}, to enforce the generated dynamics to obey fundamental physical laws. 

\am{A key direction for future research lies in developing unified, adaptive, and structure-aware representations that can seamlessly handle transitions across motion types, spatial scales, and topological changes while preserving temporal coherence and physical plausibility. Current methods often balance efficiency, fidelity, and generalization. However, since each representation offers its own benefit and shortcoming, {\em hybrid representations\/} that combine the explicit geometric hierarchy and interpretability of structured models with the flexibility and expressiveness of implicit neural representations would be an interesting direction to explore.}

\am{The role of structure remains particularly underexplored, especially how hierarchical, part-based, or physically grounded representations can enhance motion reasoning, interaction modeling, and compositionality in 4D learning.} \rz{In addition, prominent adoptions of structured representations are often found in CAD domains, with Constructive Solid Geometry (CSG), Sketch-and-Extrude, and Boundary Representations (B-Reps) offering precise geometry and compact parameterization. While many 4D applications in CAD can be envisioned, e.g., assembly or mechanical reconstruction, animation, and visualization, robot training and simulation, as well as any sub-task in the realm of 4D digital twins, a key open challenge is the {\em lack of industrial precision} by current methods, as CAD imposes stringent geometric and topological constraints across time. Making matters worse is the fact that high-quality dynamic CAD datasets are virtually nonexistent.}

\rz{Indeed, we are still faced with a general 4D dataset bottleneck. Current 4D datasets are generally small in scale with limited motion and interaction diversity, and they often lack complete, measurable geometry, or physical plausibility, resulting in models that inevitably inherit various dataset-specific biases and miss essential components or properties required to represent complete and realistic motions. There remains a pressing need to develop large-scale, standardized 4D benchmarks which encompass diverse object categories, articulated motions, and real-world dynamics.}

\rz{
To date, data-driven 4D models still frequently generate visually compelling but physically implausible dynamics. To this end, existing learning paradigms could evolve beyond reconstruction or supervision-driven training, moving toward self-supervised, causal, and physics-informed approaches 
that can infer motion, interaction, and intent directly from sparse and multi-modal input. In general, integrating semantic, material, and physical properties into 4D representations offers a promising path toward interpretable, generalizable, and application-ready models for vision, graphics, and robotics. On the other hand, disentangled representations where geometry, motion, appearance, and lighting are independently manipulable would benefit controllability and editability.}

\bibliographystyle{libs/eg-alpha-doi}  
\bibliography{ref, ref_kai}

\end{document}